\pdfoutput=1

\documentclass[11pt]{article}

\usepackage[]{ACL2023}
\usepackage{makecell}
\usepackage{times}
\usepackage{latexsym}
\usepackage{booktabs}
\usepackage{colortbl}

\definecolor{green1}{rgb}{0.19, 0.75, 0.45}
\definecolor{green2}{rgb}{0.56, 0.85, 0.68}
\definecolor{green3}{rgb}{0.78, 0.92, 0.83}
\definecolor{green4}{rgb}{0.92, 0.95, 0.93}
\definecolor{green5}{rgb}{1, 1, 1}

\usepackage[T1]{fontenc}

\usepackage{multirow}
\usepackage{multicol}
\usepackage{amsmath}
\usepackage{amssymb}
\usepackage{graphicx}
\usepackage{soul}
\usepackage{color, xcolor}

\usepackage[utf8]{inputenc}

\usepackage{microtype}

\usepackage{inconsolata}

\title{\texttt{HAUSER}: Towards Holistic and Automatic Evaluation of Simile Generation}

\author{Qianyu He\textsuperscript{\rm 1},
\textbf{Yikai Zhang}\textsuperscript{\rm 1},
\textbf{Jiaqing Liang}\textsuperscript{\rm 2}\thanks{\enspace Corresponding author.},\\
\textbf{Yuncheng Huang}\textsuperscript{\rm 1},
\textbf{Yanghua Xiao}\textsuperscript{\rm 1}\footnotemark[1],
\textbf{Yunwen Chen}\textsuperscript{\rm 3}
\\
            \textsuperscript{\rm 1}Shanghai Key Laboratory of Data Science, School of Computer Science, Fudan University\\
    \textsuperscript{\rm 2}School of Data Science, Fudan University 
    \textsuperscript{\rm 3}DataGrand Inc., Shanghai, China\\
    \{qyhe21, ykzhang22, yunchenghuang22\}@m.fudan.edu.cn, \\
    \{liangjiaqing, shawyh\}@fudan.edu.cn,
    chenyunwen@datagrand.com
    }
    
\begin{document}
\maketitle
\begin{abstract}
Similes play an imperative role in creative writing such as story and dialogue generation.
Proper evaluation metrics are like a beacon guiding the research of simile generation~(SG).
However, it remains under-explored as to what criteria should be considered, how to quantify each criterion into metrics, and whether the metrics are effective for comprehensive, efficient, and reliable SG evaluation.
To address the issues, we establish \texttt{HAUSER}, a holistic and automatic evaluation system for the SG task, which consists of five criteria from three perspectives and automatic metrics for each criterion. 
Through extensive experiments, we verify that our metrics are significantly more correlated with human ratings from each perspective compared with prior automatic metrics.
Resources of \texttt{HAUSER} are publicly available at \url{https://github.com/Abbey4799/HAUSER}.
\end{abstract}

\section{Introduction}
Similes play a vital role in human expression, making literal sentences imaginative and graspable.
For example, Robert Burns famously wrote ``\textit{My Luve is like a red, red rose}'' to metaphorically depict the beloved as being beautiful.
In this simile, ``\textit{Luve}'' (a.k.a. topic) is compared with ``\textit{red rose}'' (a.k.a. vehicle) via the implicit property ``\textit{beautiful}'' and the event ``\textit{is}''. 
Here, topic, vehicle, property, and event are four main \textit{simile components}~\cite{hanks2013lexical}.
As a figure of speech, similes have been widely used in literature and conversations~\cite{zheng2019love, chakrabarty2022s}.

\begin{figure}[t] 
    \centering
        \includegraphics[width=1.0\linewidth]{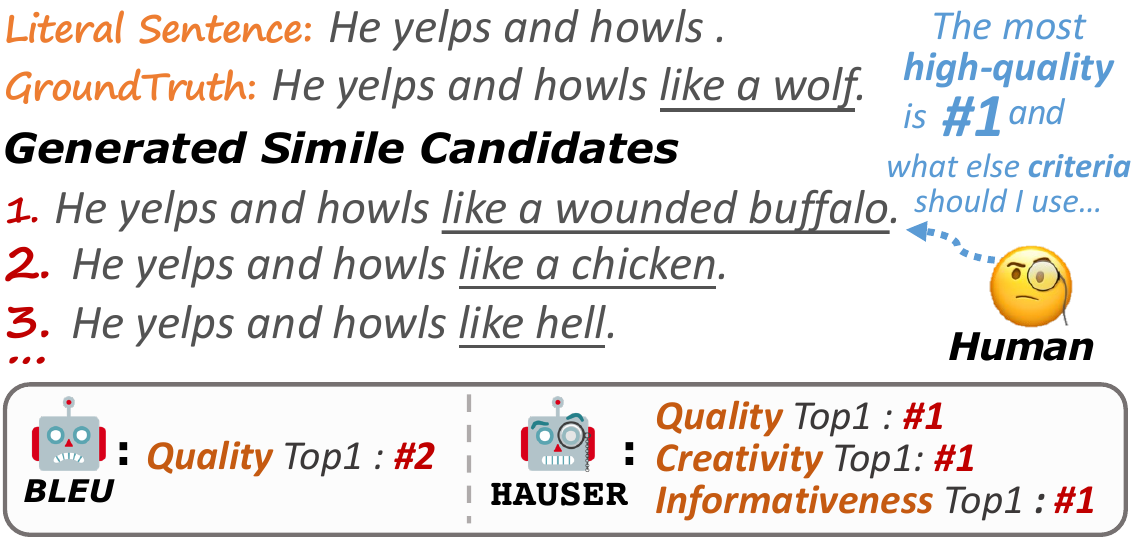} 
    \captionsetup{font={small}} 
    \caption{An example of Simile Generation~(SG) Evaluation. The commonly used automatic metric BLEU deems the second candidate as the most \textit{high-quality} one among all the generated similes, while our proposed metrics \texttt{HAUSER} deem the first candidate as the best one regarding its \textit{quality}, \textit{creativity} and \textit{informativeness}, which better correlates with human ratings and also provides more criteria for SG evaluation.}
    \label{fig:intro}
\end{figure}

Simile generation (SG) is a crucial task in natural language processing~\cite{chakrabarty2020generating, zhang2021writing, lai2022multi}, with the aim of polishing literal sentences into similes.
In Fig.~\ref{fig:intro}, the literal sentence ``\textit{He yelps and howls.}'' is polished into a simile by inserting the phrase ``\textit{like a wolf}'', resulting in ``\textit{He yelps and howls like a wolf}''.
The ability to generate similes can assist various downstream tasks, such as making the generations more imaginative in story or poet generation task~\cite{tartakovsky2018simple, chakrabarty2022s} and the generated response more human-like in dialogue generation task~\cite{zheng2019love}.

\begin{table*}[t]
\small
\centering
    \resizebox{1.0\textwidth}{!}{
\begin{tabular}{cp{1.5cm}p{3.9cm}p{9.7cm}}
  \toprule
\multicolumn{2}{c}{\textbf{Criterion}}                       & \multicolumn{1}{c}{\textbf{Literal Sentence}}                                        & \multicolumn{1}{c}{\textbf{Example Simile Candidates}}                                                                  \\
\midrule
\multirow{3}{*}{\textbf{Quality}} & \makecell[c]{\textbf{Relevance}}             & Some raindrops struck the roof, window and ran down its panes.            & Some raindrops struck the roof, window and ran down its panes (\textbf{like tears} | like arrows).             \\
        \cmidrule{2-4}
                         & \begin{tabular}[c]{@{}c@{}}\textbf{Logical}\\ \textbf{Consistency}\end{tabular}  & \makecell*[l]{ Stefan moved, every movement \\easy and precisely controlled. }                & \makecell*[l]{Stefan moved (like lightning | \textbf{like a dancer}), every movement easy and  \\ precisely controlled.}           \\
         \cmidrule{2-4}
                         & \begin{tabular}[c]{@{}c@{}}\textbf{Sentiment}\\ \textbf{Consistency}\end{tabular}& \makecell*[l]{The idea resounded throughout \\ the land.}                                     & \makecell*[l]{The idea resounded (like an earthquake | \textbf{like a thunderous wave}) throughout \\ the land.}                 \\
    \midrule
\multicolumn{2}{c}{\textbf{Creativity}}                   & He possessed a power of sarcasm which could scorch. &  He possessed a power of sarcasm which could scorch (\textbf{like vitriol} | like fire). \\
\midrule
\multicolumn{2}{c}{\textbf{Informativeness}}              & They gleamed.                                                               & They gleamed (like the eyes of a cat | \textbf{like the eyes of an angry cat}).\\                       \bottomrule
\end{tabular}
}

    \caption{Examples of our criteria for Simile Generation~(SG) Evaluation. We design five criteria from three perspectives. The vehicles of the better simile candidates given by each criterion are highlighted in bold.}
    \label{tab:metric_exp}
\end{table*}

Automatic evaluation is critical for the SG task since it enables efficient, systematic, and scalable comparisons between models in general~\cite{celikyilmaz2020evaluation}.
However, existing studies are inadequate for effective SG evaluation.
Task-agnostic automatic metrics~\cite{papineni2002bleu,zhang2019bertscore, li2016diversity} are widely adopted for SG evaluation~\cite{zhang2021writing,lai2022multi}, which have several limitations:
(1) The simile components should receive more attention than other words during SG evaluation (e.g. ``\textit{he}'' and ``\textit{wolf}'' in Fig. \ref{fig:intro}), while there are no automatic metrics that consider the key components.
(2) The SG task is open-ended, allowing for multiple plausible generations for the same input~\cite{chakrabarty2020generating} (e.g. the howling man can be compared to ``\textit{wolf}'', ``\textit{buffalo}'', or ``\textit{tiger}'' in Fig.~\ref{fig:intro}).
Hence, the metrics based on word overlap with a few references are inadequate to accurately measure the overall quality of generated similes.
As shown in Fig.~\ref{fig:intro}, the commonly used metric BLEU deems the second candidate as the highest quality, as it has more overlapped words with the only referenced groundtruth, while human deems the first candidate as the most coherent one.
(3) The existing metrics are inadequate to provide fine-grained and comprehensive SG evaluation, considering that the creative generation tasks have distinct criteria for desired generations~\cite{celikyilmaz2020evaluation}, such as novelty and complexity for story generation~\cite{chhun2022human} and logical consistency for dialogue generation~\cite{pang2020towards}.

\begin{figure*}[!htb] 
    \centering
        \includegraphics[width=1.0\linewidth]{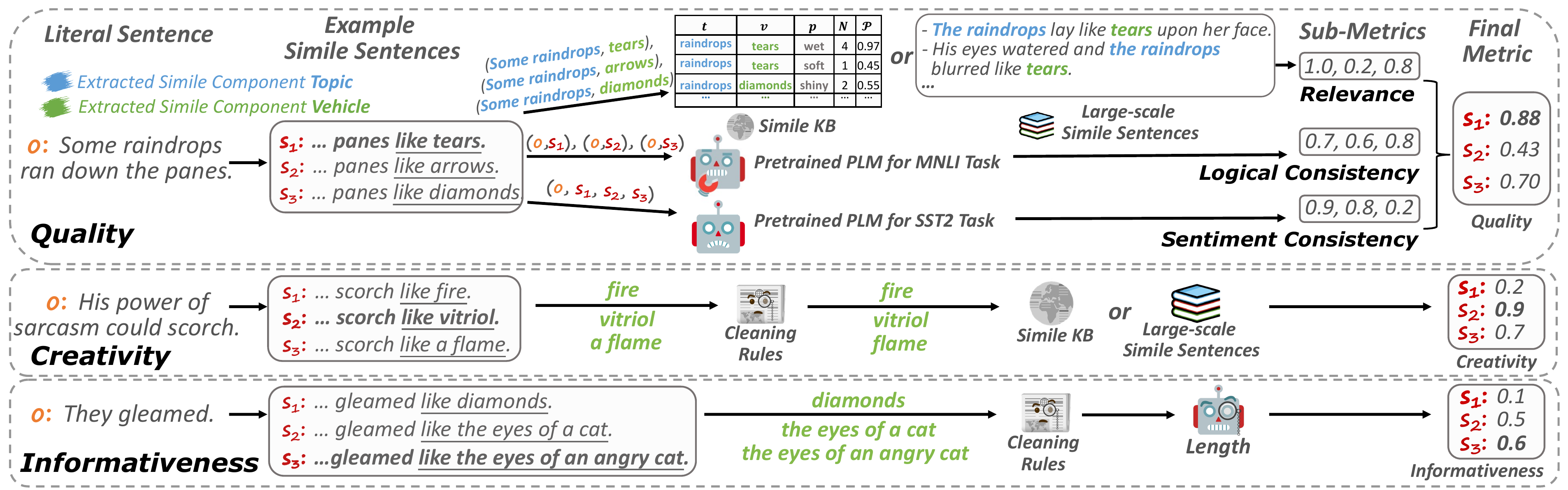} 
    \captionsetup{font={small}} 
    \caption{The framework of our automatic metrics design. We design the automatic metric for each criterion in Tab. \ref{tab:metric_exp}.}
    \label{fig:framework}
\end{figure*}

However, establishing a comprehensive, efficient, and reliable evaluation system for SG is non-trivial, which raises three main concerns:
(1) What criteria should be adopted to evaluate the SG task in a comprehensive and non-redundant fashion?
(2) How to quantify each criterion into a metric thus enabling efficient and objective SG evaluation, given that the human evaluation of creative generation task is not only time-consuming but also subjective and blurred~\cite{niculae2014brighter, celikyilmaz2020evaluation}?
(3) Whether the proposed metrics are effective in providing useful scores to guide actual improvements in the real-world application of the SG model?
In this paper, we establish \texttt{HAUSER}, a \textbf{\underline{H}}olistic and \textbf{\underline{AU}}tomatic evaluation system for \textbf{\underline{S}}imile g\textbf{\underline{E}}ne\textbf{\underline{R}}ation task, consisting of five criteria (Tab.~\ref{tab:metric_exp}):
(1) The \textit{relevance} between topic and vehicle, as the foundation of a simile is to compare the two via their shared properties~\cite{paul1970figurative}.
(2) The \textit{logical consistency} between the literal sentence and generated simile, since the aim of SG task is to polish the original sentence without altering its semantics~\cite{tversky1977features}.
(3) The \textit{sentiment consistency} between the literal sentence and generated simile, since similes generally transmit certain sentiment polarity~\cite{qadir2015learning}.
(4,5) The \textit{creativity} and \textit{informativeness} of the simile, since novel similes or those with richer content can enhance the literary experience~\cite{jones2006roosters, roncero2015semantic, addison2001so}.
Overall, these five criteria can be categorized into three perspectives: \textit{quality} (which considers relevance, logical, and sentiment consistency jointly), \textit{creativity}, and \textit{informativeness}.
We further quantify each criterion into automatic metrics (Fig. \ref{fig:framework}) and prove their effectiveness through extensive experiments.

To the best of our knowledge, we are the first to systematically investigate the automatic evaluation of the SG task.
To summarize, our contributions are mainly three-fold:
(1) We establish a holistic and automatic evaluation system for the SG task, consisting of five criteria based on linguistic theories, facilitating both human and automatic evaluation of this task.
(2) We design automatic metrics for each criterion, facilitating efficient and objective comparisons between SG models.
(3) We conduct extensive experiments to verify that our metrics are significantly more correlated with human ratings than prior metrics.


\section{Related Work}
\subsection{Simile Generation Task}
There are two primary forms of the simile generation~(SG) task: simile triplet completion and literal sentence polishing.
For simile triplet completion, a model receives simile components, topic and property, and is required to generate the vehicle~\cite{roncero2015semantic, zheng2019love, chen2022probing, he2022maps}.
For literal sentence polishing, a model receives a literal sentence and is expected to convert it into similes~\cite{zhang2021writing, stowe2020metaphoric,chakrabarty2020generating, lai2022multi}.
We focus on the latter.
However, prior works mainly adopt task-agnostic automatic metrics to evaluate the SG task, raising concern as to whether the claimed improvements are comprehensive and reliable.

\subsection{Automatic Evaluation for NLG Systems}
Existing automatic metrics for Natural Language Generation~(NLG) evaluation can be categorized into task-agnostic and task-specific metrics.
Task-agnostic metrics can be applied to various NLG tasks, which generally focus on the coherence of generations~\cite{papineni2002bleu, zhang2019bertscore}, including n-gram-based metrics~\cite{papineni2002bleu, lin2004rouge, denkowski2014meteor} and embedding-based metrics~\cite{zhang2019bertscore, zhao2019moverscore}.
There are also many metrics for evaluating the diversity of generations~\cite{li2016diversity, zhu2018texygen, tevet2021evaluating}.
Task-specific metrics are proposed to evaluate NLG systems on specific tasks~\cite{tao2018ruber, dhingra2019handling, ren2020codebleu}.
Specifically, various works systematically study the evaluation of the creative generation task~\cite{pang2020towards, tevet2021evaluating, chhun2022human}.
Different from these works, we revisit SG evaluation, propose holistic criteria based on linguistic theories, and design effective automatic metrics for it.

\section{\texttt{HAUSER} for SG evaluation}
We establish \texttt{HAUSER}, a holistic and automatic evaluation system for SG evaluation, containing five criteria from three perspectives, and further design automatic metrics for each criterion (Fig. \ref{fig:framework}).

\subsection{Quality} \label{sec:quality}
We measure the overall quality of generated similes using three criteria: \textit{relevance}, \textit{logical consistency}, \textit{sentiment consistency}.
The key simile components - topic and vehicle - should be relevant, as the foundation of a simile is to compare the two via their shared properties (\textit{relevance})~\cite{paul1970figurative}.
In Tab.~\ref{tab:metric_exp}, comparing ``\textit{raindrops}'' to ``\textit{tears}'' is more coherent than to ``\textit{arrows}''.
Additionally, the generated simile should remain logically consistent with the original sentence (\textit{logical consistency}), as the SG task aims to polish the plain text without changing its semantics~\cite{tversky1977features}.
In Tab.~\ref{tab:metric_exp}, comparing ``\textit{Stefan}'' to ``\textit{dancer}'' better depicts his controlled and easy movement than to ``\textit{lightning}''.
Furthermore, as similes generally transmit certain sentiment polarity~\cite{qadir2015learning}, the generated simile should enhance the sentiment polarity of the original sentence (\textit{sentiment consistency}).
In Tab.~\ref{tab:metric_exp}, the vehicle ``\textit{thunderous wave}'' enhances the positive polarity of the original sentence, while the vehicle ``\textit{earthquake}'' brings a negative sentiment polarity in opposition to the original sentence.

\subsubsection{Relevance}
For the \textit{relevance} score, if the components of one simile are relevant, they tend to co-occur in simile sentences~\cite{xiao2016meta4meaning, he2022maps} and possess shared properties~\cite{paul1970figurative, tversky1977features}.
Hence, obtaining the relevance score requires large-scale simile sentences as references, as well as knowledge about the properties (adjectives) of each simile component.
For a simile $s$, the relevance score is defined as follows:

\vspace{-1.5mm}
\begin{small}
\begin{align}
		& r = \frac{1}{m_p}\sum_{(t,v)\in s}\sum_{e \in \Gamma (t,v)}P_e(t,v),
\end{align} 
\end{small}
where there are $m_p$ topic-vehicle pairs extracted from simile $s$, each denoted as $(t, v)$$\footnote{All the simile components in our work are extracted and cleaned using rules from~\cite{he2022maps} which determines the optimal semantics a component should carry, e.g., ``a kid in a candy store'' instead of just ``a kid''.}$.
$\Gamma (t,v)$ is the set of similes containing $(t, v)$ as simile components, each denoted as $e$.
$P_e(t,v)$ is the probability that the simile components $(t, v)$ share properties in the context of the simile sentence $e$.

An effective way to obtain the frequency information $\Gamma (t,v)$ and property knowledge $P_e(t,v)$ is to utilize the large-scale probabilistic simile knowledge base MAPS-KB~\cite{he2022maps}, which contains millions of simile triplets in the form of (\textit{topic}, \textit{property}, \textit{vehicle}), along with frequency and two probabilistic metrics to model each triplet\footnote{More details of MAPS-KB is provided in Appx.~\ref{sec:mapskb}}.
Specifically, the probabilistic metric \textit{Plausibility} is calculated based on the confidence score of the simile instance (\textit{topic}, \textit{property}, \textit{vehicle}, \textit{simile sentence}) supporting the triplet, indicating the probability that the topic and vehicle share the property.
The relevance score $r$ can be calculated as follow:

\vspace{-2.6mm}
\begin{small}
\begin{align}
		&r = \frac{1}{m_p} \sum_{(t,v) \in s}\sum_{(t, p, v) \in \mathcal{G}_{(t,v)}} n(t,p,v)\cdot \mathcal{P}(t,p,v),
\end{align} 
\end{small}
where $\mathcal{G}_{(t,v)}$ is the set of triplets ($t$, $p$ ,$v$) containing the ($t$, $v$) pair in MAPS-KB, with $p$ referring to the property.
$n$ and $\mathcal{P}$ are the metrics provided by MAPS-KB, where $n$ and $\mathcal{P}$ denote the frequency and the plausibility of the triplet respectively.

It is noticed that the metric is not coupled with MAPS-KB, as the frequency information can be obtained by referencing a large set of simile sentences and the property knowledge can be contained via other knowledge bases.
More methods are beyond the scope of this paper.
However, we additionally provide a method to approximate the relevance score.
If we assume the probability that the simile components $(t, v)$ share properties in each sentence is 1, the relevance score can be approximated as:

\vspace{-2.0mm}
 \begin{small}
\begin{align}
		&r \approx \frac{1}{m_p}\sum_{(t,v)\in s}n(t,v),
\end{align} 
 \end{small}
\vspace{-0.5mm}
where $n(t,v)$ denotes the number of samples that contain the simile components $(t, v)$ in large-scale simile sentences.
We discuss the effects of the referenced dataset size in Sec. \ref{sec:effect_size}.


\subsubsection{Logical Consistency}
The literal sentence and the generated simile that are logically inconsistent generally exhibit contradictory logic.
Hence, for a generated simile, we input the <literal text($l$), simile($s$)> sentence pair into existing pre-trained Multi-Genre Natural Language Inference (MNLI) model$\footnote{We use the checkpoint of the model (roberta-base\_mnli\_bc) that achieves the SOTA performance on the GLUE~\cite{wang2018glue} MNLI dataset at the time of submission, according to https://paperswithcode.com/sota/text-classification-on-glue-mnli. }$, which determines the relation between them is \textit{entailment}, \textit{neutral}, or \textit{contradiction}.
The logical consistency score $c_l$ of this simile is defined as follows~\cite{pang2020towards}:

\vspace{-3mm}
\begin{small}
\begin{align}
		& c_l = 1 - P(h_{<l,s>} = c),
\end{align} 
\end{small}
where $P(h_{<l,s>} = c)$ represents the probability that the model predicts the relation of the sentence pair $<l,s>$ to be \textit{contradiction} (denoted as $c$).

\subsubsection{Sentiment Consistency}
 Better similes tend to enhance the sentiment polarity of the original sentence~\cite{qadir2015learning}.
Hence, we first apply the model fine-tuned on the GLUE SST-2 dataset$\footnote{We apply the checkpoint of the model (distilbert-base-uncased-finetuned-sst-2-english) with the most download times on the GLUE SST-2 dataset at the time of submission, according to https://huggingface.co/models.}$ to classify each simile as being either \textit{positive} or \textit{negative}. 
Then, the sentiment consistency score $c_s$ is defined as follows:

\vspace{-3mm}
\begin{small}
\begin{align}
		& c_s = P(h_s = a) - P(h_l = a),
\end{align} 
\end{small}
where $a$ is the sentiment polarity of the literal sentence (\textit{positive} or \textit{negative}) predicted by the model.
$P(h_s = a)$ and $P(h_l = a)$ denote the probabilities that the model predicts the sentiment polarity of the simile $s$ and the literal sentence $l$ to be $a$, respectively.



It is noticed that different <topic, vehicle> pairs within a sentence may have distinct sentiment polarities, such as <\textit{She}, \textit{scared rabbit}> and <\textit{I}, \textit{bird}> in the simile ``\textit{If she escapes like a scared rabbit, I will fly like a bird to catch her.}''.
Directly inputting text containing multiple topic-vehicle pairs into the sentiment classification model will result in inferior performance.
Therefore, for each simile, only the text from the beginning up to the first \textit{vehicle} is input into the model (i.e. ``\textit{If she escapes like a scared rabbit}'' in the given example), and for each literal sentence, the text from the beginning up to the first \textit{event} (i.e. ``\textit{If she escapes}'' in the given example) is input into the model.

\subsubsection{Combination}
Since the aim of the SG task is to polish the plain text, the quality of similes generated from different texts can not be compared.
Therefore, the normalized score among the simile candidates for each original text is utilized.
Suppose there are $m$ simile candidates $\mathcal{S} = \{s_1, s_2, ..., s_m\}$ for the literal text $l$, the original relevance scores of $\mathcal{R}$ is $\mathcal{R} = \{r_1, r_2, ..., r_m\}$ respectively.
The normalized relevance score ${r}'_i$ of $s_i$ is formulated as follows:

\vspace{-1.8mm}
\begin{small}
\begin{align}
		& {r}'_i = \frac{r_i - min(\mathcal{R})}{max(\mathcal{R}) - min(\mathcal{R})},
\end{align} 
\end{small}
which ranges from 0 to 1.
Then, the normalized logical and sentiment consistency score ${c}'_{li}$, ${c}'_{si}$ for each simile $s_i$ are obtained in the same manner\footnote{If all the relevance scores $r_i$ in $\mathcal{R}$ are the same, the normalized relevance scores $r'_i$ in $\mathcal{R}'$ are set to 0.5 uniformly.}.

Finally, the \textit{quality} for simile $s_i$ is defined as the weighted combination of three parts as follows: 

\vspace{-1mm}
\begin{small}
\begin{align}
		&\mathcal{Q}_i = \alpha \cdot r'_i + \beta \cdot {c}'_{li} + \gamma \cdot {c}'_{si},
\end{align} 
\end{small}
where $\alpha$, $\beta$, and $\gamma$ are hyperparameters.

\subsection{Creativity} \label{sec:creativity}
Creative similes can provide a better literary experience~\cite{jones2006roosters}.
In Tab.~\ref{tab:metric_exp}, comparing ``\textit{sarcasm}'' to ``\textit{vitriol}'' is less common than to ``\textit{fire}'', yet it better conveys the intensity of a person's sarcasm.
Hence, we design \textit{creativity} score.

Previous studies mainly evaluate the creativity of text generation tasks via human evaluation~\cite{sai2022survey}, since measuring the creativity of open-ended text is a relatively difficult task~\cite{celikyilmaz2020evaluation}.
Although there have been many works evaluating the diversity of open-ended text generation~\cite{li2016diversity, zhu2018texygen, tevet2021evaluating}, these metrics are not suitable for measuring the creativity of the text.
Because the \textit{diversity} metrics take a set of generated text as input and output one score, while a \textit{creativity} metric is required to measure each text individually and output a set of corresponding scores.


Different from other open-ended generation tasks, the components of the generated similes enable us to evaluate creativity automatically.
According to linguists, the creativity of a simile is determined by vehicles~\cite{pierce2008roles, roncero2015semantic}.
Intuitively, the generated simile may be less creative if its extracted topic-vehicle pair co-occurs frequently, or if many topics are compared to its vehicle in the corpus.
Therefore, we adopt large-scale corpora as references when designing our creativity metric.
The creativity score of $s$ is calculated as follows:

\vspace{-1.2mm}
\begin{small}
\begin{align}
		&\mathcal{C}_{i} = -log(\frac{1}{m_v}\sum_{v \in s} N_v + 1),
\end{align} 
\end{small}
\vspace{-0.5mm}
where there are $m_v$ vehicles extracted from the simile $s$, each denoted as $v$. $N_v$ denotes the frequency of the vehicles appearing in the similes in the corpora.
The log transformation aims to reduce the influence of extreme values.

An effective way to obtain the adequate frequency information $N_v$ is to utilize the million-scale simile knowledge base MAPS-KB, where the $N_v$ can be defined as follows:

\vspace{-1.8mm}
\begin{small}
\begin{align}
		&N_v = \sum_{(t, p, v) \in \mathcal{G}_{v}} n(t,p,v),
\end{align} 
\end{small}
$\mathcal{G}_{v}$ is the set of triplets containing the vehicle $v$ in MAPS-KB, $n$ denotes the frequency of the triplet.

It is noticed that the metric is not coupled with MAPS-KB, as $N_v$ can also be obtained by counting the samples containing the vehicle $v$ in large-scale simile sentences.
The method of obtaining the simile sentences is beyond the scope of this paper. 
Nevertheless, we discuss the effects of the referenced dataset size in Sec. \ref{sec:effect_size_creativity}.

\subsection{Informativeness}
The vehicle with richer content can create a more impact and vivid impression\cite{addison2001so}.
In the example from Tab.~\ref{tab:metric_exp}, the addition of the word ``\textit{angry}'' makes the similes more expressive.
Therefore, we design the metric \textit{informativeness} to measure the content richness of the vehicles.

Intuitively, the more words a vehicle contains, the richer its content will be.
Hence, for a given simile $s$, we adopt the average length of the extracted vehicles to be the \textit{informativeness} score\footnote{Different from the \textit{quality} metric, we do not use a normalized score for \textit{creativity} and \textit{informativeness}, since they mainly depend on the generated vehicles, rather than the original text.}~\cite{chakrabarty2020generating, zhang2021writing}, defined as $\mathcal{I}_i = \frac{1}{m_v}\sum_{v \in s}\text{len}(v)$, where there are $m_v$ vehicles extracted from simile $s$.

\section{\texttt{HAUSER} Analysis}
In this section, we conduct experiments to verify the effectiveness of our automatic metrics.

\begin{figure*}[!htbp] 
    \centering
        \includegraphics[width=0.9\linewidth]{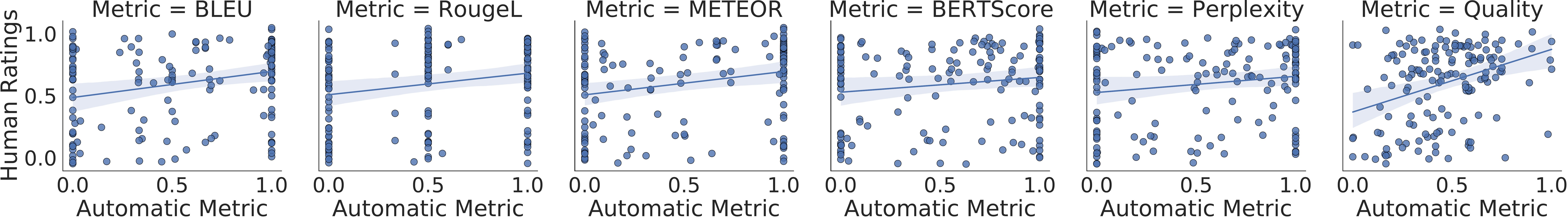} 
    \captionsetup{font={small}} 
    \caption{Correlation between automatic metrics and human ratings when evaluating quality. 
    Here, BLEU2, Rouge2, and $\text{BERTScore}_{\text{large}}$ are presented since they perform the best in their respective category.
    To avoid overlapping points, random jitters sampled from $\mathcal{N}(0, 0.05^2)$ were added to human ratings after fitting the regression.}
    \label{fig:regression_quality}
    \vspace{-1.5mm}
\end{figure*}

\subsection{Experiment Setup}  \label{sec:setup}

\subsubsection{Simile Generation} 
The existing datasets for the SG task are either Chinese~\cite{zhang2021writing}, limited to the simile triplet completion~\cite{roncero2015semantic,chen2022probing}, or having all vehicles located at end of the sentence~\cite{chakrabarty2022s, lai2022multi}, which are not practical for English simile generation in a real-world application.
To bridge the gap, we construct a large-scale English dataset for SG task based on simile sentences from ~\cite{he2022maps}, which contains 524k simile sentences labeled with topic and vehicle.
The output decoder target is the simile sentence $s$ and the input encoder source is $s$ rewritten to drop the comparator ``\textit{like}'' and the vehicle.
For example, given $s$ =  ``\textit{\textit{The idea resounded \underline{like a thunderclap} throughout the land.}}'', the encoder source would be ``\textit{The idea resounded throughout the land.}''.
In particular, we remove the simile sentences whose event is a linking verb (e.g. \textit{be}, \textit{seem}, \textit{turn}) as they would be meaningless after the vehicle is removed.
The final train, validation and test sets contain 139k, 2.5k, and 2.5k sentence pairs, respectively.

Based on our constructed dataset, we fine-tune a pre-trained sequence-to-sequence model, BART~\cite{lewis2020bart}, for the SG task, which has been demonstrated to be an effective framework for various figurative language generation~\cite{zhang2021mover, chakrabarty2022s, he2022maps, lai2022multi}.
The experiments are run on RTX3090 GPU and the implementation of BART is based on the HuggingFace Transformers\footnote{https://github.com/huggingface/transformers/}.
The experiments are run with a batch size of 16, a max sequence length of 128, and a learning rate of 4e-5 for 10 epochs.

\subsubsection{Evaluation Dataset Construction} 
Firstly, we randomly sample 50 literal sentences from the test set and adopt the trained SG model to generate five candidates for each one.
Then, for each perspective, three raters are asked to rate each simile from 1 to 5, where 1 denotes the worst and 5 denotes the best\footnote{The details about human ratings, including the instructions provided to raters and examples of human ratings are provided in Appx.~\ref{sec:human_ratings}.}. 
Since evaluating the quality of generated similes is subjective and blurred~\cite{niculae2014brighter}, we remove the simile-literal sentence pairs if (1) raters argue that the pairs lack context and are difficult to rate (e.g. ``\textit{Nobody can shoot.}'') or (2) some raters rate them as low quality (quality score of 1-2), while others rate them as high quality (scores of 4-5)~\cite{niculae2014brighter}.
Moreover, we measure the inter-rater agreement by holding out the ratings of one rater at a time, calculating the correlations with the average of the other rater's ratings, and finally calculating the average or maximum of all the held-out correlations (denoted as ``\textit{Mean}'' and ``\textit{Max}'', respectively).
The inter-rater agreement before and after applying the filtering strategies is shown in Tab.~\ref{tab:inter-rater}.
Overall, the final inter-rater agreement ensures the reliability of our evaluation of automatic metrics and the filtering strategies improve the inter-rater agreement generally.
We finally get 150 simile candidates generated from 44 literal sentences.

\begin{table}[t]
 \centering
 \resizebox{0.4\textwidth}{!}{
\begin{tabular}{cccccc}
\toprule  
\multirow{2}{*}{\textbf{Setting}} & \multirow{2}{*}{\textbf{Metric}} & \multicolumn{2}{c}{\textbf{Pearson}} & \multicolumn{2}{c}{\textbf{Spearman}} \\
                        \cmidrule{3-6}
                         &                         & \textbf{Mean}         & \textbf{Max}          & \textbf{Mean}          & \textbf{Max}          \\
  \midrule  
\multirow{3}{*}{\textbf{Before}}  & Quality                 & 0.573        & \textbf{0.626}        & \textbf{0.542 }        & \textbf{0.595 }       \\
                         & Creativity              & \textbf{0.537 }       & 0.671        & 0.550         & 0.678        \\
                         & Informativeness         & 0.833        & 0.857        & 0.799         & 0.816        \\
  \midrule  
\multirow{3}{*}{\textbf{After}}   & Quality                 & 0.812        & 0.833        & 0.735         & 0.759        \\
                         & Creativity              & 0.551        & 0.643        & 0.568         & 0.650        \\
                         & Informativeness         & 0.848        & 0.893        & 0.817         & 0.841       \\
\bottomrule 
\end{tabular}
}
\caption{The inter-rater agreement before and after applying the removal strategies.
Bold numbers are the worst results, indicating that the raters are quite divided on this metric.
}
 \label{tab:inter-rater}
\end{table}

\subsection{Results}
\subsubsection{Quality}
We compare our \textit{quality} metric with the following automatic metrics\footnote{These metrics are normalized among simile candidates for a literal sentence, since the quality score between the similes generated from different literal sentences can not be compared.
Please refer to Appx.~\ref{sec:appendix_evaluation} for the implementation of them.}:
(1) \textbf{BLEU}~\cite{papineni2002bleu} calculates the precision of n-gram matches,
(2) \textbf{RougeL}~\cite{lin2004rouge} is a recall-oriented metric,
(3) \textbf{METEOR}~\cite{denkowski2014meteor} proposes a set of linguistic rules to compare the hypothesis with the reference, 
(4) \textbf{BERTScore}~\cite{zhang2019bertscore} calculates the cosine similarity between the BERT embeddings,
(5) \textbf{Perplexity}~\cite{pang2020towards} measures the proximity of a language model, the inverse of which is utilized.

\textbf{Correlations with Human Ratings.}
Tab. \ref{tab:correlation_quality} shows the correlation coefficients between automatic metrics and human ratings.
Firstly, our metrics are significantly more correlated with human ratings than prior automatic metrics.
Moreover, all the sentence-level metrics, which consider the semantics of the entire sentence, perform worse than almost all the n-gram-level metrics, which compare the n-grams between the hypothesis and the reference, which reveals that simile components need to be specifically considered during SG evaluation.

\begin{table}[t] 
    \small
    \centering
    \resizebox{0.30\textwidth}{!}{
    
        \begin{tabular}{c|c|c}
        \toprule  
       \textbf{Metrics}        & \textbf{Pearson} & \textbf{Spearman}   \\
        \midrule  
        \multicolumn{3}{c}{\textbf{N-gram-level Metrics}} \\
         \midrule  
        BLEU1   &   0.229    & 0.218   \\
        BLEU2   &   \underline{0.255}    &  0.208  \\
        BLEU3   & 0.193    &  0.172  \\
        BLEU4   &  \textit{ 0.159 }   &  \textit{0.140}   \\
        Rouge1 &    0.185    &  0.176   \\
        Rouge2 &   0.210    & 0.190   \\
        RougeL &   0.173    & \textit{ 0.152}   \\
        METEOR &   0.234    &  \underline{0.233}   \\
          \midrule  
        \multicolumn{3}{c}{\textbf{Sentence-level Metrics}} \\
         \midrule  
        $\text{BERTS}_{\text{base}}$ &  \textit{0.107}      & \textit{0.075}   \\
       $\text{BERTS}_{\text{large}}$ &  \textit{0.143}      & \textit{0.120}   \\
        Perplexity  &  \textit{0.157}    & \textit{0.120}    \\
          \midrule  
        \multicolumn{3}{c}{\textbf{\texttt{HAUSER}}} \\
         \midrule  
            $\text{Quality}$   &  \textbf{0.320}(+6.5\%)      &  \textbf{0.292}(+5.9\%)  \\
            \midrule  
                    $-\text{relevance}$ &   0.206      &  0.194   \\
        $-\text{consistency}_l$  &    0.259     &  0.217    \\
        $-\text{consistency}_s$  &    0.307    &    0.265   \\ 
        \bottomrule 
        \end{tabular}
        }
       \captionsetup{font={small}} 
        \caption{Correlation between automatic metrics and human ratings when evaluating quality. All measures with p-value $> 0.05$ are italicized. Bold numbers are the best results. The second best results are marked by ``$\underline{\ \ \ \ \ \ }$''. ``$-$'' denotes the removal of the sub-metric.}    
    \label{tab:correlation_quality}
    
\end{table}

\begin{table}[t] 
    \scriptsize
    \centering
     \resizebox{0.45\textwidth}{!}{
        \begin{tabular}{c|c|c|c|c|c}
        \toprule  
       \textbf{Metrics}        & \textbf{HR@1} & \textbf{HR@3}  & \textbf{nDCG@1} & \textbf{nDCG@3} & \textbf{MRR} \\
        \midrule  
        \multicolumn{6}{c}{\textbf{N-gram-level Metrics}} \\
         \midrule  
        BLEU1   & \underline{0.429}    &  \textbf{0.857} & 0.893 & \textbf{0.945} & 0.662 \\
        BLEU2   &  0.314   &  0.838 & 0.892 & 0.936 & 0.600\\
        BLEU3   & 0.286   & 0.838  & 0.859 & 0.924 & 0.648\\
        BLEU4   & 0.286    &0.838  & 0.882 & 0.929 & 0.581\\
        Rouge1   & 0.400    &\underline{0.848}  & \underline{0.907} & \underline{0.941} & 0.655\\
        Rouge2   & 0.400    &\underline{0.848}  & 0.905 & 0.937 & 0.650\\
        RougeL &  \underline{0.429}     & \underline{0.848} & 0.901 & 0.937 & \underline{0.670} \\
        METEOR &  0.286  & \textbf{0.857} & 0.884  & 0.936 & 0.589 \\
          \midrule  
        \multicolumn{6}{c}{\textbf{Sentence-level Metrics}} \\
         \midrule  
        $\text{BERTS}_{\text{base}}$ &   0.314 &  0.829 & 0.870  & 0.934 & 0.585 \\
        $\text{BERTS}_{\text{large}}$&   0.257  &  0.838 & 0.895 & 0.939 & 0.570 \\
        Perplexity  & 0.257 &  0.810 &0.898 & 0.940 & 0.549 \\
          \midrule  
                \multicolumn{6}{c}{\textbf{\texttt{HAUSER}}} \\
         \midrule  
        Quality & \textbf{0.457}     & \underline{0.848} &\textbf{0.915}&  0.937 & \textbf{0.688}\\
        \bottomrule 
        \end{tabular}
        }
       \captionsetup{font={small}} 
        \caption{Comparison of automatic metrics ranking and human ranking when evaluating quality.}    
    \label{tab:recom_quality}
\end{table}

According to the visualized correlation result in Fig.~\ref{fig:regression_quality}, datapoints from prior automatic metrics tend to scatter at 0 or 1, while the datapoints from our metric are distributed closer to the fitter line, proving that our metric can better measure the quality.



\textbf{Recommendation Task.} 
We compare the rankings given by automatic metrics with human rankings\footnote{We remove the literal sentences with fewer than three valid simile candidates in this task, as they are too simple to rank. 
We finally get 134 sentences from 35 literal sentences. }.
We adopt the following metrics: Hit Ratio at rank K (\textbf{HR@K}(K=1,3)), Normalized Discounted Cumulative Gain at rank K (\textbf{NDCG@K}(K=1,3))\footnote{The formulated NDCG@K in our setting is provided in Appx.~\ref{sec:appendix_ndcg}, with the optimal ranking being human rankings.}, and Mean Reciprocal Rank (\textbf{MRR}).
From Tab. \ref{tab:recom_quality}, our metric achieves significant improvement compared to other metrics, indicating that our metric can yield more accurate rankings for quality.
Also, the n-gram-level metrics generally outperform sentence-level metrics, which is consistent with the result in Tab. \ref{tab:correlation_quality}.





\textbf{Ablation Study.}
To investigate the importance of different sub-metrics in \textit{quality} metric, we compare the correlation between \textit{quality} metric and human ratings after removing each sub-metric individually.
From Tab.~\ref{tab:correlation_quality}, the removal of any sub-metric leads to a decline in performance, which proves the effectiveness of each sub-metric.
Among three components, the removal of the \textit{relevance} results in the largest performance drop, which reveals that \textit{relevance} is the most important sub-metric.

       

\begin{figure}[t]

\centering
\includegraphics[width=0.48\textwidth]{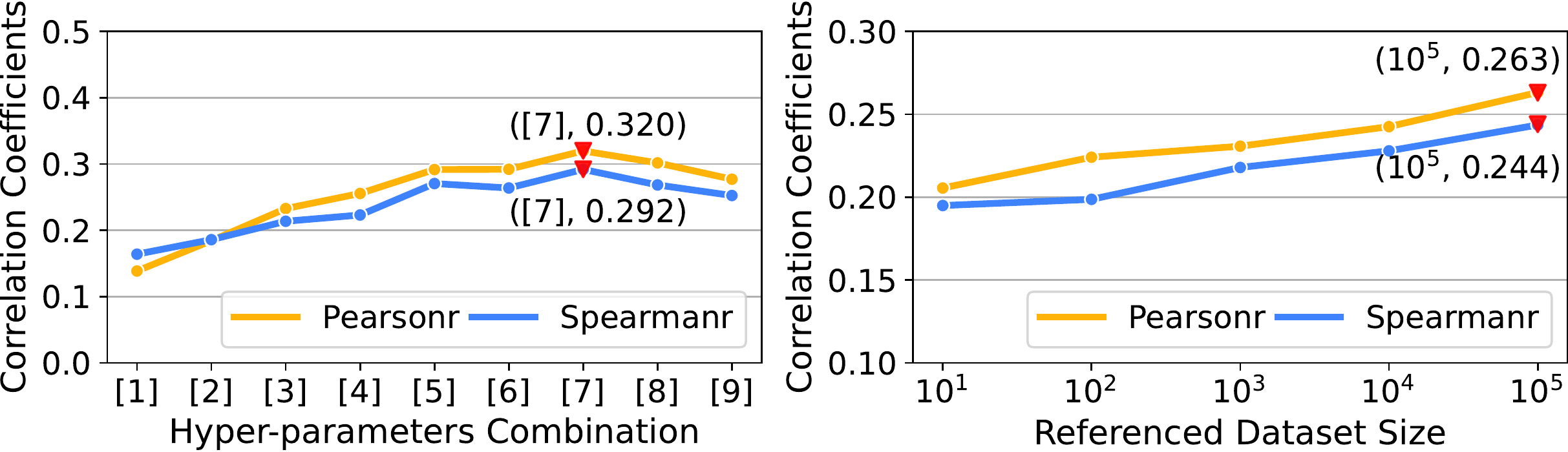}  

\caption{Correlation between \textit{quality} metric and human ratings with different hyperparameters (left) and different referenced corpus size (right).}
\label{fig:hyper+corpus_size_quality}

 \vspace{-2mm}
\end{figure}

\textbf{The Effects of Hyperparameters.}
Since different sub-metrics have varying levels of importance, we study the correlation results when gradually increasing the weight of \textit{relevance} component and decreasing the weight of \textit{sentiment consistency} component (as in Tab. \ref{tab:hyper}).
From Fig. \ref{fig:hyper+corpus_size_quality} (left), increasing the weight of the \textit{relevance} component consistently results in improved performance, peaking at the combination [7]($\alpha, \beta, \gamma = 3/6, 2/6, 1/6$), before eventually causing a decline in performance.
This reveals that although \textit{relevance} is the most important sub-metric, too much weight on it can be detrimental.

\begin{table}[t] 
    \small
    \centering
      \resizebox{0.25\textwidth}{!}{
        \begin{tabular}{c|c}
        \toprule  
       \textbf{Combination}        & \textbf{$\alpha,\beta,\gamma$}   \\
        \midrule  
        $\left [ 1 \right ]$  &   1/12, 1/12, 5/6      \\
        $\left [ 2 \right ]$ &   1/6, 1/6, 4/6      \\
        $\left [ 3 \right ]$  &   1/6, 2/6, 3/6      \\
        $\left [ 4 \right ]$  &   1/6, 3/6, 2/6      \\
        $\left [ 5 \right ]$ &    2/6, 2/6, 2/6    \\
        $\left [ 6 \right ]$  &   2/6, 3/6, 1/6      \\
        $\left [ 7 \right ]$  &   3/6, 2/6, 1/6      \\
        $\left [ 8 \right ]$  &   4/6, 1/6, 1/6      \\
        $\left [ 9 \right ]$  &   5/6, 1/12, 1/12      \\
        \bottomrule 
        \end{tabular}
        }
       \captionsetup{font={small}} 
        \caption{The setting of each hyperparameters combination for the \textit{quality} metric. The result is shown in Fig. \ref{fig:hyper+corpus_size_quality} (left).}    
    \label{tab:hyper}
\end{table}










\textbf{The Effects of Referenced Dataset Size.}  \label{sec:effect_size}
We sample different numbers of simile sentences from~\cite{he2022maps} as references for relevance score and study the correlation between the \textit{quality} metric and human ratings\footnote{The results are averaged over three random seeds.}.
From Fig.~\ref{fig:hyper+corpus_size_quality} (right)\footnote{The best hyper-parameter combination is applied.}, correlations grow linearly with exponential growth in referenced dataset size, indicating that using datasets larger than 100k will improve the correlation coefficients.
Moreover, the performance at the peak surpasses the prior automatic metrics, proving the effectiveness of our approximation method.



\subsubsection{Creativity}
We compare our \textit{creativity} metric with the following automatic metrics:
(1) \textbf{Perplexity} which is often utilized to measure diversity as well~\cite{tevet2021evaluating},
(2) \textbf{Self-BLEU}~\cite{zhu2018texygen} calculates the BLEU score of each generation against all other generations as references,
(3) \textbf{Distinct n-grams(Dist)}~\cite{tevet2021evaluating}, which is the fraction of distinct n-grams from all possible n-grams across all generations.

\begin{table}[t] 
    \small
    \centering
      \resizebox{0.31\textwidth}{!}{
        \begin{tabular}{c|c|c}
        \toprule  
       \textbf{Metrics}        & \textbf{Pearson} & \textbf{Spearman}   \\
        \midrule  
        \multicolumn{3}{c}{\textbf{Prior Diversity Metrics}} \\
        \midrule  
         Perplexity   &  \textit{0.088} & \textit{0.041} \\
         Self-BLEU3   &  \textit{0.118}      &  \textit{0.076} \\
         Self-BLEU4   &  0.196      &  \textit{0.175} \\
         Self-BLEU5   &  0.128      &  \textit{0.077} \\
         Dist1   &   0.278  & 0.311  \\
         Dist2   &  \underline{0.319}      & 0.369 \\
         Dist3   &   0.299   &  \underline{0.379} \\
           \midrule  
                 \multicolumn{3}{c}{\textbf{\texttt{HAUSER}}} \\
           \midrule  
        $\text{Creativty}$ &    \textbf{0.592}(+27.3\%)     &  \textbf{0.645}(+26.6\%)    \\
        \midrule  
         $-\text{log}$ &    0.394    &    0.571   \\ 
        \bottomrule 
        \end{tabular}
        }
       \captionsetup{font={small}} 
        \caption{Correlation between metrics and human ratings when evaluating creativity. All measures with p-value $> 0.05$ are italicized. ``$-$\text{log}'' denotes the removal of log transformation.}    
    \label{tab:correlation_creativity}
     \vspace{-2.5mm}
\end{table}

\begin{figure*}[t]
    \centering
        \includegraphics[width=0.8\linewidth]{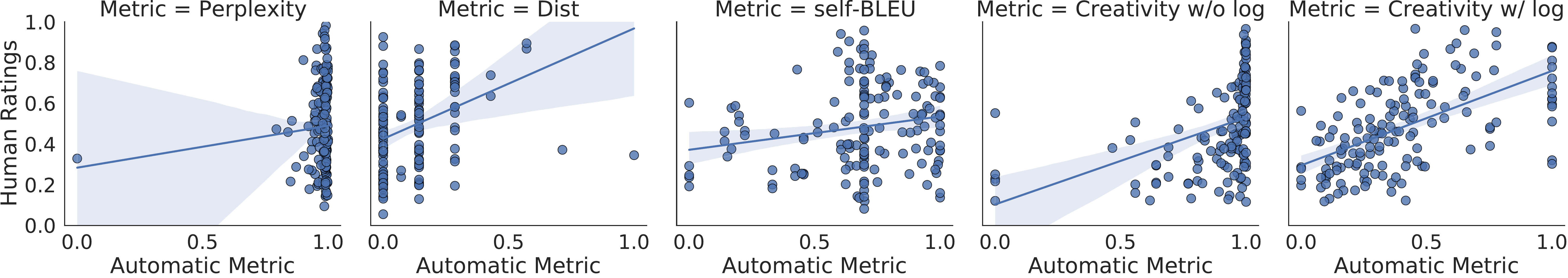} 
    \captionsetup{font={small}} 
    \caption{Correlation between automatic metrics and human ratings when evaluating creativity. Here, Self-BLEU4 and Dist2, which perform the best in their respective category in Tab.~\ref{tab:correlation_creativity}, are presented. ``w/o log'' and ``w/ log'' denotes whether the log transformation is applied or not. }
    \label{fig:regression_creativity}
     \vspace{-2mm}
\end{figure*}

\textbf{Correlations with Human Ratings.}
From Tab.~\ref{tab:correlation_creativity}, our metric \textit{creativity} is significantly more correlated with human evaluation scores compared with prior diversity metrics.
According to the visualized correlation result in Fig.~\ref{fig:regression_creativity}, the prior diversity metrics have either wide confidence intervals (Perplexity, Dist) or scattered datapoints (self-BLEU), whereas our creativity metrics exhibit stronger linear correlation and narrower confidence intervals (Creativty w/ Log), implying higher reliability.

\textbf{Recommendation Task. }
We compare the rankings given by automatic metrics with human rankings.
According to Tab. \ref{tab:recom_creativity}, our creativity metric outperforms prior automatic metrics, which proves our metric can better measure the creativity of simile candidates given a literal sentence, which is consistent with the results in Tab. \ref{tab:correlation_creativity}.

\begin{table}[t] 
    \scriptsize
    \centering
    \resizebox{0.46\textwidth}{!}{
        \begin{tabular}{c|c|c|c|c|c}
        \toprule  
       \textbf{Metrics}        & \textbf{HR@1} & \textbf{HR@3}  & \textbf{nDCG@1} & \textbf{nDCG@3} & \textbf{MRR} \\
        \midrule  

         \multicolumn{6}{c}{\textbf{Prior Diversity Metrics}} \\
           \midrule  
            Perplexity & 0.314 & 0.800& 0.800& 0.903 & 0.566 \\
        Self-BLEU3 & 0.257 & 0.771 & 0.765 & 0.892 & 0.520 \\
        Self-BLEU4 & 0.257 & 0.762 & 0.756 & 0.889 & 0.518 \\
        Self-BLEU5 & 0.229 & 0.762 & 0.751 & 0.882 & 0.504 \\
        Dist1 & 0.486 & 0.800 & 0.862 & 0.927 & 0.671  \\
        Dist2 & \underline{0.571} & 0.810 & \underline{0.893} & \underline{0.939} & \underline{0.737}  \\
        Dist3 & 0.543 & \underline{0.838} & 0.877 & 0.938 & 0.725  \\
        
               \midrule  
                 \multicolumn{6}{c}{\textbf{\texttt{HAUSER}}} \\
           \midrule  
           
         $\text{Creativty}$  & \textbf{0.629} &\textbf{0.914} & \textbf{0.944} & \textbf{0.976} & \textbf{0.784}\\
         
        \bottomrule 
        \end{tabular}
        }
       \captionsetup{font={small}} 
        \caption{Comparison of automatic metrics ranking and human ranking when evaluating creativity.}    
    \label{tab:recom_creativity}
     \vspace{-2mm}
\end{table}

\textbf{Ablation Study.}
According to Tab. \ref{tab:correlation_creativity}, removing the log transformation leads to significant performance drops.
According to the visualized correlation result in Fig. \ref{fig:regression_creativity}, the datapoints are distributed closer to the fitter line and exhibit narrower confidence intervals after applying the log transformation, which further proves that log transformation is essential for our creativity metric. 


\textbf{The Effects of Referenced Dataset Size.}
\label{sec:effect_size_creativity}
According to Fig.~\ref{fig:corpus_size+correlation_info} (left), the correlation coefficients increase continuously and eventually converge as the number of referenced sentences increases.
Moreover, the performance after convergence is comparable to that given by the \textit{creativity} metric based on the simile KB.
The trend reveals that our metric referencing 10k similes can achieve a promising correlation with human ratings.

\vspace{-1mm}

\subsubsection{Informativeness}
\begin{figure}[t] 
\begin{minipage}[t]{0.58\linewidth} 
\centering
\includegraphics[width=1.7in, height=1.1in]{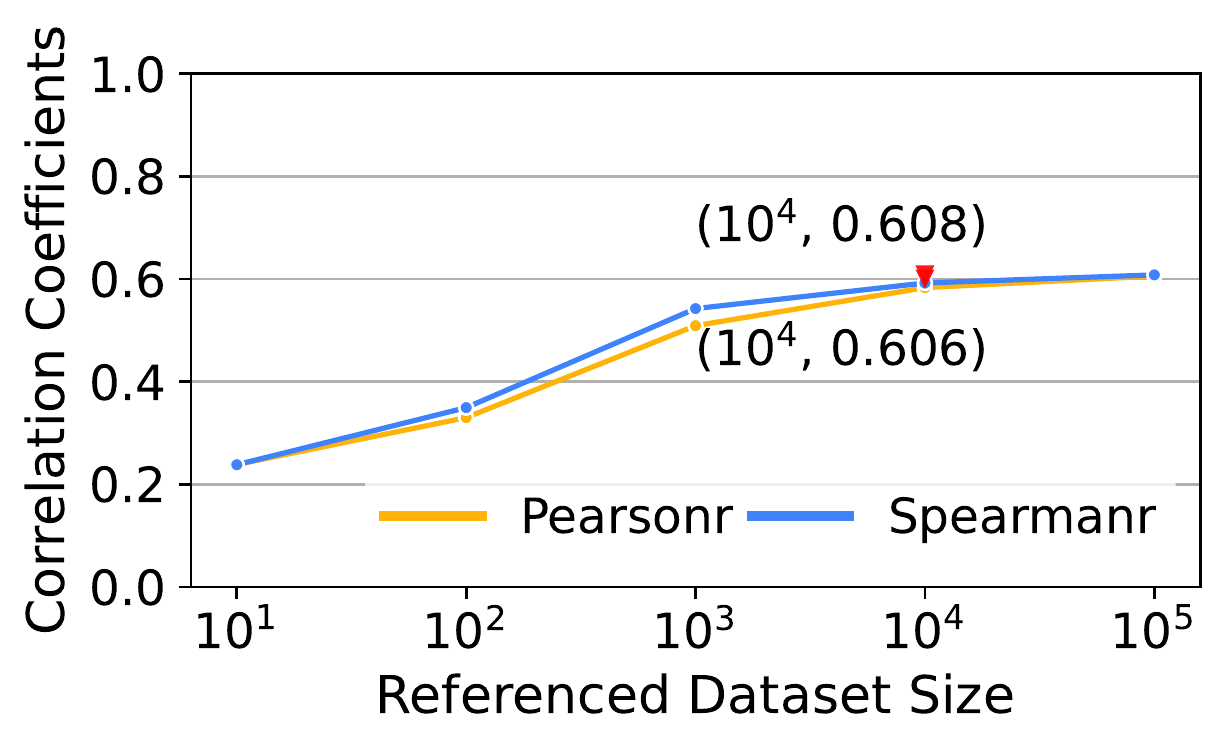}  
\label{fig1} 
\end{minipage}%
\begin{minipage}[t]{0.4\linewidth}
\centering
\includegraphics[width=1.2in, height=1.2in]{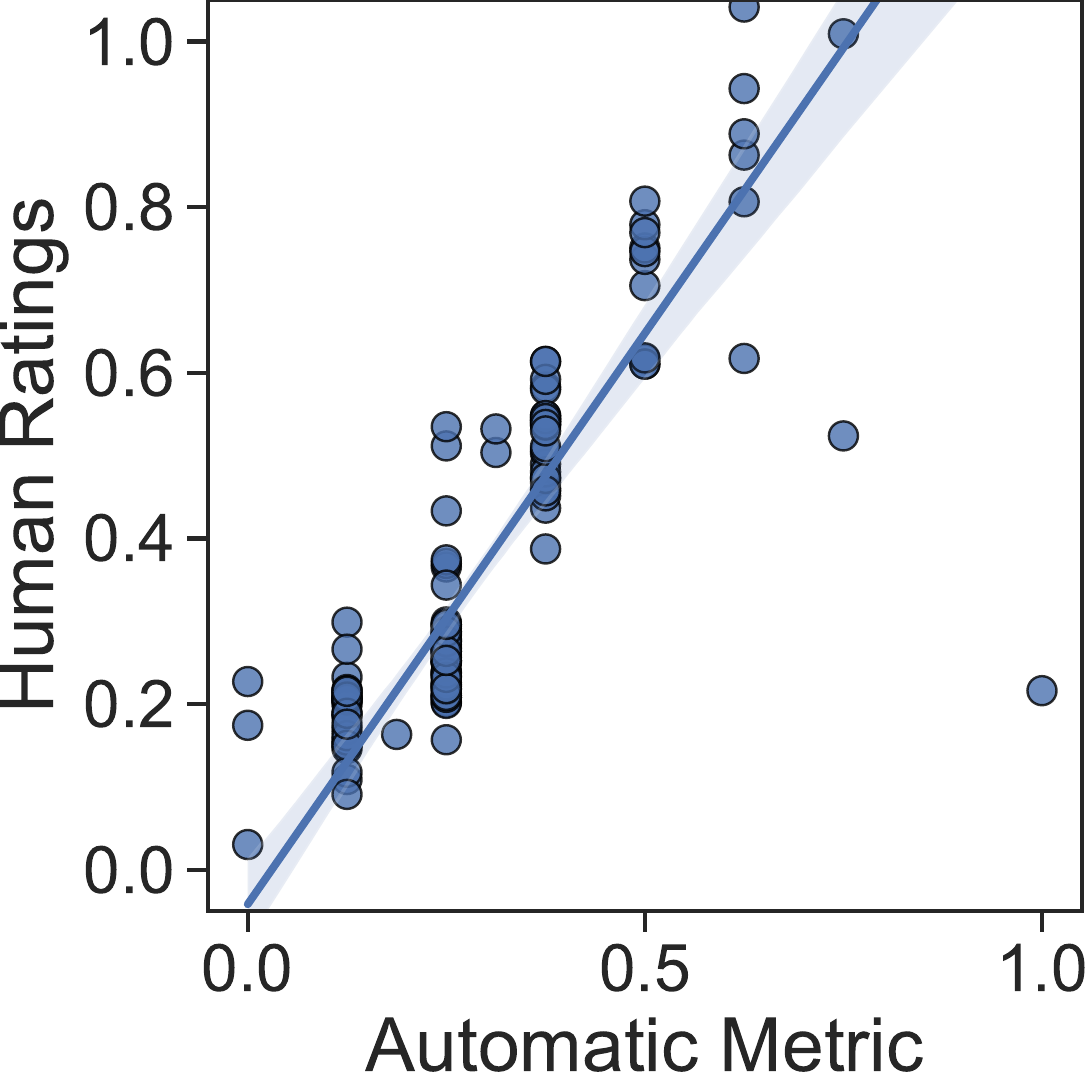} 
\label{fig2}
\end{minipage}%
\vspace{-4mm}
\caption{Correlation between \textit{creativity} metric and human ratings with varying referenced corpus size (left), and correlation between \textit{informativeness} metric and human ratings (right). }
 \label{fig:corpus_size+correlation_info}
\end{figure}

The Pearson and Spearman correlation coefficients between our \textit{informativeness} metric and human ratings are 0.798 and 0.882, respectively.
According to Fig. \ref{fig:corpus_size+correlation_info}~(right), the strong linear correlation between the metric and human ratings proves that our \textit{informativeness} metric is simple yet quite effective.

\subsubsection{Relation between Metrics}

\begin{figure}[t] 
\begin{minipage}[t]{0.32\linewidth} 
\centering
\includegraphics[width=0.9in, height=0.9in]{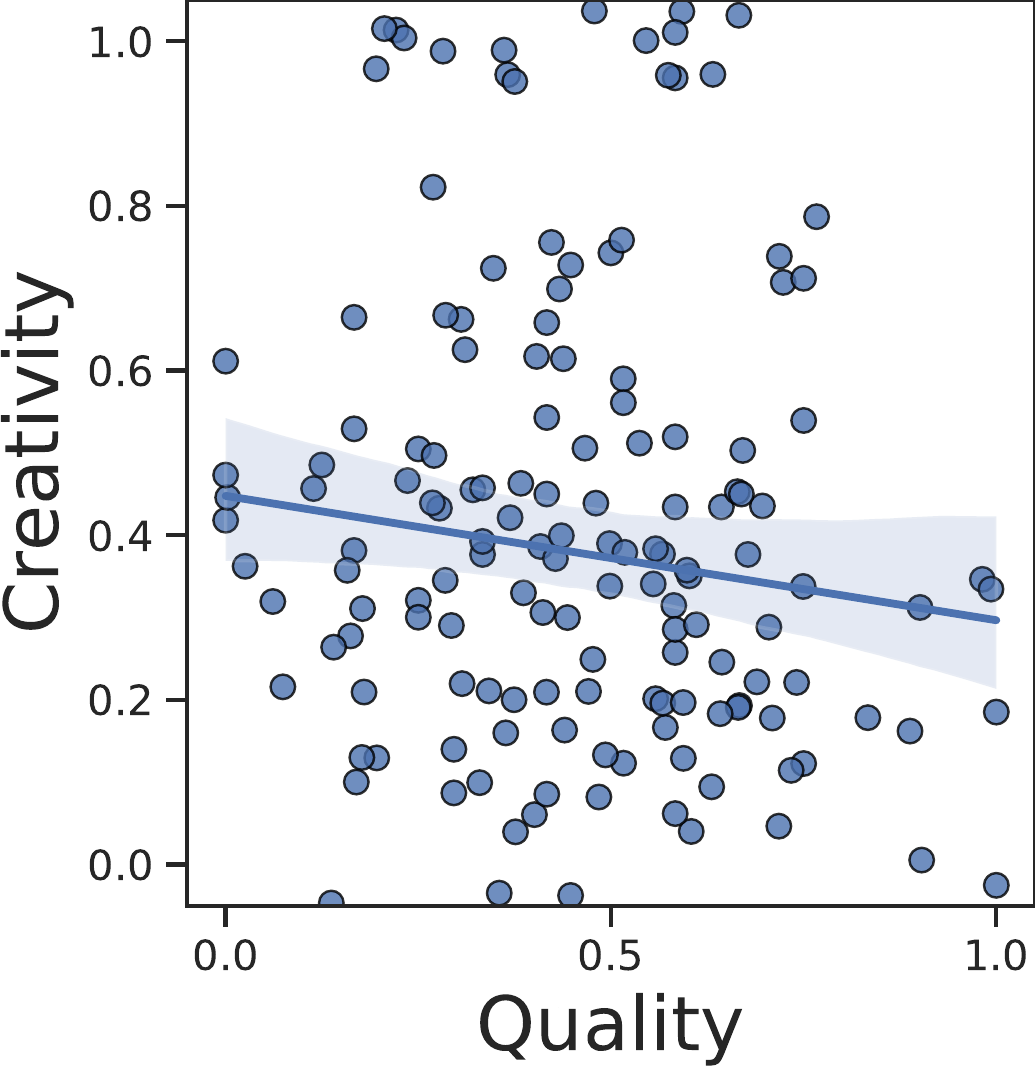} 
\label{fig1} 
\end{minipage}%
\begin{minipage}[t]{0.32\linewidth}
\centering
\includegraphics[width=0.9in, height=0.9in]{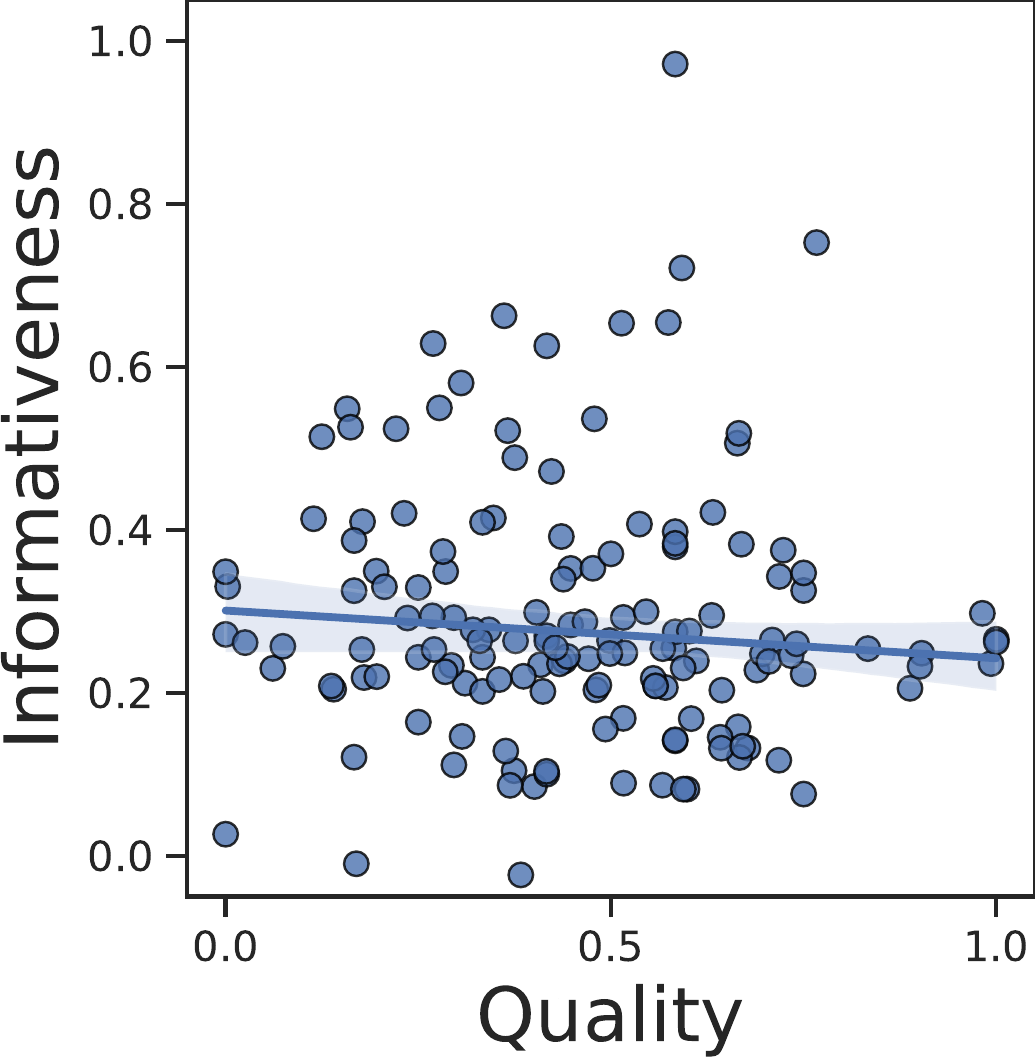}
\label{fig2}
\end{minipage}%
\begin{minipage}[t]{0.32\linewidth}
\centering
\includegraphics[width=0.9in, height=0.9in]{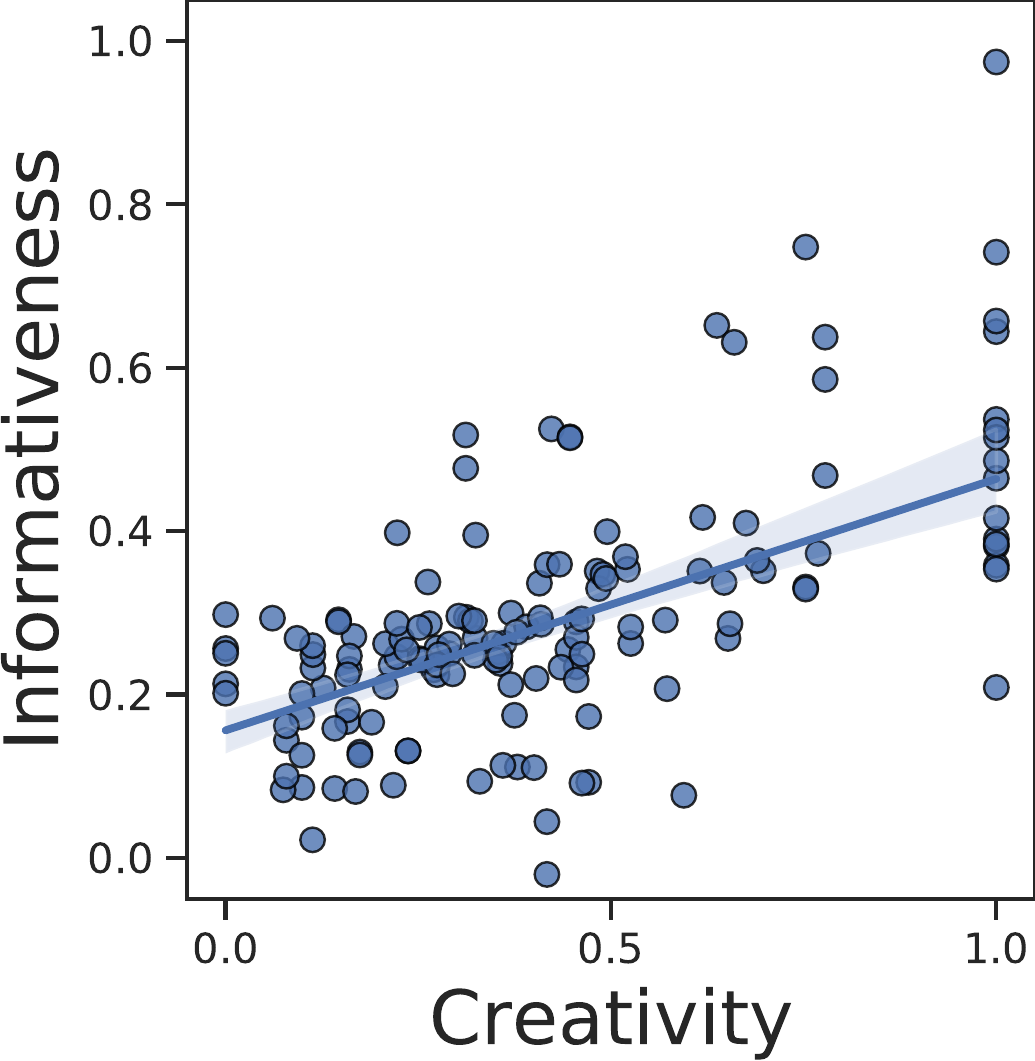}

\label{fig3}
\end{minipage}
\caption{The pair-wise correlations between the metrics.}
 \label{fig:regression_pair}
\end{figure}

 
We present pair-wise correlations between the three automatic metrics in Tab. \ref{tab:correlation_pair} and also visualize them in Fig.~\ref{fig:regression_pair}.
Among the three metrics, creativity correlates with informativeness moderately, mainly because shorter vehicles tend to be less creative than longer ones.
The correlations of all other pair-wise metrics are relatively weak.
Thus, it is evident that the three metrics are independent of each other and it is necessary to measure each one of them to obtain a holistic view of SG evaluation.

\begin{table}[!htbp] 
    \small
    \centering
      \resizebox{0.37\textwidth}{!}{
        \begin{tabular}{c|c|c}
        \toprule  
       \textbf{Metrics}        & \textbf{Pearson} & \textbf{Spearman}   \\
        \midrule  
        Quality \& Creativity & \textit{-0.116} &  \textit{-0.130} \\
        Quality \& Informativeness & \textit{-0.040} & \textit{-0.118}  \\
        Creativity \& Informativeness & 0.652 & 0.635  \\
        \bottomrule 
        \end{tabular}
        }
       \captionsetup{font={small}} 
        \caption{The pair-wise correlations between our automatic metrics. All measures with p-value $> 0.05$ are italicized.}    
    \label{tab:correlation_pair}
\end{table}

\vspace{-6mm}
\section{\texttt{HAUSER} Application} \label{sec:application}

We perform a case study to prove that our designed automatic metrics are effective for various methods.
Here, we apply our metrics to a retrieval method~\cite{zhang2021writing} (denoted as \textbf{BM25}), which utilizes the 20 context words around the insertion position given by groundtruth to retrieve the 5 most similar samples based on the BM25 ranking score from the training set, and adopts the vehicles from these samples to be those of simile candidates.
This method ensures the diversity of generated similes.
The method introduced in Sec. \ref{sec:setup} is denoted as \textbf{Ours}.
Given the candidates generated by each method, we rerank them using a weighted combination of quality, creativity, and informativeness rankings obtained by \texttt{HAUSER}, with a ratio of 2:2:1.

From Tab. \ref{tab:application} in Appendix, the candidates generated by various methods can be more correlated with human rankings after being ranked by our metrics, thus proving the generality of our metrics.
It is noticed that the insertion position for \textbf{BM25} is provided by the groundtruth, while the insertion position for \textbf{Ours} is predicted by the model, thus proving the effectiveness of our generation method.

\section{Conclusion}
In this work, we systematically investigate the evaluation of the Simile Generation~(SG) task.
We establish a holistic and automatic evaluation system for the SG task, containing five criteria from three perspectives, and propose holistic automatic metrics for each criterion.
Extensive experiments verify the effectiveness of our metrics.

\section*{Acknowledgements}
This research is funded by the Science and Technology Commission of Shanghai Municipality Grant (No. 22511105902), Shanghai Municipal Science and Technology Major Project (No. 2021SHZDZX0103), National Natural Science Foundation of China (No. 62102095).

\section*{Limitations}
We analyze the limitations of our work as follows.
Firstly, although applying a million-scale simile knowledge base or large-scale simile sentences as reference makes our designed metric significantly more correlated with humans than prior reference-based metrics (e.g. BLEU, Rouge, BERTScore), our metrics are still reference-based and rely on the quality and scale of referenced data. 
We have discussed the effect of referenced dataset size in our paper and will design reference-free metrics to further complement our metrics in future work.
Additionally, since our metrics utilize a million-scale simile knowledge base or large-scale simile sentences as references, the efficiency of our method is slightly lower than the automatic metrics based on a few references.
Nevertheless, this limitation does not prevent our metrics from performing systematic and scalable comparisons between SG models.

\section*{Ethical Considerations}
We provide details of our work to address potential ethical considerations.
In our work, we propose holistic and automatic metrics for SG evaluation and construct an evaluation dataset to verify their effectiveness (Sec.~\ref{sec:setup}).
All the data sources used in our evaluation dataset are publicly available.
The details about human ratings, such as the instructions provided to raters, are provided in Appx.~\ref{sec:human_ratings}.
In our case study (Sec.~\ref{sec:application}), the human rankings are discussed by three raters.
We protect the privacy rights of raters.
All raters have been paid above the local minimum wage and consented to use the evaluation dataset for research purposes covered in our paper.
Our work does not raise any ethical considerations regarding potential risks and does not involve the research of human subjects.

\bibliography{anthology,custom}

\begin{thebibliography}{38}
\expandafter\ifx\csname natexlab\endcsname\relax\def\natexlab#1{#1}\fi

\bibitem[{Addison(2001)}]{addison2001so}
Catherine Addison. 2001.
\newblock “so stretched out huge in length”: Reading the extended simile.
\newblock \emph{Style}, 35(3):498--516.

\bibitem[{Bosselut et~al.(2019)Bosselut, Rashkin, Sap, Malaviya, Celikyilmaz,
  and Choi}]{bosselut2019comet}
Antoine Bosselut, Hannah Rashkin, Maarten Sap, Chaitanya Malaviya, Asli
  Celikyilmaz, and Yejin Choi. 2019.
\newblock Comet: Commonsense transformers for automatic knowledge graph
  construction.
\newblock \emph{arXiv preprint arXiv:1906.05317}.

\bibitem[{Celikyilmaz et~al.(2020)Celikyilmaz, Clark, and
  Gao}]{celikyilmaz2020evaluation}
Asli Celikyilmaz, Elizabeth Clark, and Jianfeng Gao. 2020.
\newblock Evaluation of text generation: A survey.
\newblock \emph{arXiv preprint arXiv:2006.14799}.

\bibitem[{Chakrabarty et~al.(2022)Chakrabarty, Choi, and
  Shwartz}]{chakrabarty2022s}
Tuhin Chakrabarty, Yejin Choi, and Vered Shwartz. 2022.
\newblock It’s not rocket science: Interpreting figurative language in
  narratives.
\newblock \emph{Transactions of the Association for Computational Linguistics},
  10:589--606.

\bibitem[{Chakrabarty et~al.(2020)Chakrabarty, Muresan, and
  Peng}]{chakrabarty2020generating}
Tuhin Chakrabarty, Smaranda Muresan, and Nanyun Peng. 2020.
\newblock Generating similes effortlessly like a pro: A style transfer approach
  for simile generation.
\newblock In \emph{Proceedings of the 2020 Conference on Empirical Methods in
  Natural Language Processing (EMNLP)}, pages 6455--6469.

\bibitem[{Chen et~al.(2022)Chen, Chang, Zhang, Pu, Chen, Zhang, Xi, Chen, and
  Su}]{chen2022probing}
Weijie Chen, Yongzhu Chang, Rongsheng Zhang, Jiashu Pu, Guandan Chen, Le~Zhang,
  Yadong Xi, Yijiang Chen, and Chang Su. 2022.
\newblock Probing simile knowledge from pre-trained language models.
\newblock In \emph{Proceedings of the 60th Annual Meeting of the Association
  for Computational Linguistics (Volume 1: Long Papers)}, pages 5875--5887.

\bibitem[{Chhun et~al.(2022)Chhun, Colombo, Suchanek, and
  Clavel}]{chhun2022human}
Cyril Chhun, Pierre Colombo, Fabian~M Suchanek, and Chlo{\'e} Clavel. 2022.
\newblock Of human criteria and automatic metrics: A benchmark of the
  evaluation of story generation.
\newblock In \emph{29th International Conference on Computational Linguistics
  (COLING 2022)}.

\bibitem[{Denkowski and Lavie(2014)}]{denkowski2014meteor}
Michael Denkowski and Alon Lavie. 2014.
\newblock Meteor universal: Language specific translation evaluation for any
  target language.
\newblock In \emph{Proceedings of the ninth workshop on statistical machine
  translation}, pages 376--380.

\bibitem[{Dhingra et~al.(2019)Dhingra, Faruqui, Parikh, Chang, Das, and
  Cohen}]{dhingra2019handling}
Bhuwan Dhingra, Manaal Faruqui, Ankur Parikh, Ming-Wei Chang, Dipanjan Das, and
  William Cohen. 2019.
\newblock Handling divergent reference texts when evaluating table-to-text
  generation.
\newblock In \emph{Proceedings of the 57th Annual Meeting of the Association
  for Computational Linguistics}, pages 4884--4895.

\bibitem[{Hanks(2013)}]{hanks2013lexical}
Patrick Hanks. 2013.
\newblock \emph{Lexical analysis: Norms and exploitations}.
\newblock Mit Press.

\bibitem[{He et~al.(2022)He, Wang, Liang, and Xiao}]{he2022maps}
Qianyu He, Xintao Wang, Jiaqing Liang, and Yanghua Xiao. 2022.
\newblock Maps-kb: A million-scale probabilistic simile knowledge base.
\newblock \emph{arXiv preprint arXiv:2212.05254}.

\bibitem[{Jones and Estes(2006)}]{jones2006roosters}
Lara~L Jones and Zachary Estes. 2006.
\newblock Roosters, robins, and alarm clocks: Aptness and conventionality in
  metaphor comprehension.
\newblock \emph{Journal of Memory and Language}, 55(1):18--32.

\bibitem[{Lai and Nissim(2022)}]{lai2022multi}
Huiyuan Lai and Malvina Nissim. 2022.
\newblock Multi-figurative language generation.
\newblock In \emph{Proceedings of the 29th International Conference on
  Computational Linguistics}, pages 5939--5954.

\bibitem[{Lewis et~al.(2020)Lewis, Liu, Goyal, Ghazvininejad, Mohamed, Levy,
  Stoyanov, and Zettlemoyer}]{lewis2020bart}
Mike Lewis, Yinhan Liu, Naman Goyal, Marjan Ghazvininejad, Abdelrahman Mohamed,
  Omer Levy, Veselin Stoyanov, and Luke Zettlemoyer. 2020.
\newblock Bart: Denoising sequence-to-sequence pre-training for natural
  language generation, translation, and comprehension.
\newblock In \emph{Proceedings of the 58th Annual Meeting of the Association
  for Computational Linguistics}, pages 7871--7880.

\bibitem[{Li et~al.(2016)Li, Galley, Brockett, Gao, and
  Dolan}]{li2016diversity}
Jiwei Li, Michel Galley, Chris Brockett, Jianfeng Gao, and William~B Dolan.
  2016.
\newblock A diversity-promoting objective function for neural conversation
  models.
\newblock In \emph{Proceedings of the 2016 Conference of the North American
  Chapter of the Association for Computational Linguistics: Human Language
  Technologies}, pages 110--119.

\bibitem[{Lin(2004)}]{lin2004rouge}
Chin-Yew Lin. 2004.
\newblock Rouge: A package for automatic evaluation of summaries.
\newblock In \emph{Text summarization branches out}, pages 74--81.

\bibitem[{Niculae and Danescu-Niculescu-Mizil(2014)}]{niculae2014brighter}
Vlad Niculae and Cristian Danescu-Niculescu-Mizil. 2014.
\newblock Brighter than gold: Figurative language in user generated
  comparisons.
\newblock In \emph{Proceedings of the 2014 conference on empirical methods in
  natural language processing (EMNLP)}, pages 2008--2018.

\bibitem[{Pang et~al.(2020)Pang, Nijkamp, Han, Zhou, Liu, and
  Tu}]{pang2020towards}
Bo~Pang, Erik Nijkamp, Wenjuan Han, Linqi Zhou, Yixian Liu, and Kewei Tu. 2020.
\newblock Towards holistic and automatic evaluation of open-domain dialogue
  generation.
\newblock In \emph{Proceedings of the 58th Annual Meeting of the Association
  for Computational Linguistics}, pages 3619--3629.

\bibitem[{Papineni et~al.(2002)Papineni, Roukos, Ward, and
  Zhu}]{papineni2002bleu}
Kishore Papineni, Salim Roukos, Todd Ward, and Wei-Jing Zhu. 2002.
\newblock Bleu: a method for automatic evaluation of machine translation.
\newblock In \emph{Proceedings of the 40th annual meeting of the Association
  for Computational Linguistics}, pages 311--318.

\bibitem[{Paul(1970)}]{paul1970figurative}
Anthony~M Paul. 1970.
\newblock Figurative language.
\newblock \emph{Philosophy \& Rhetoric}, pages 225--248.

\bibitem[{Pierce and Chiappe(2008)}]{pierce2008roles}
Russell~S Pierce and Dan~L Chiappe. 2008.
\newblock The roles of aptness, conventionality, and working memory in the
  production of metaphors and similes.
\newblock \emph{Metaphor and symbol}, 24(1):1--19.

\bibitem[{Qadir et~al.(2015)Qadir, Riloff, and Walker}]{qadir2015learning}
Ashequl Qadir, Ellen Riloff, and Marilyn Walker. 2015.
\newblock Learning to recognize affective polarity in similes.
\newblock In \emph{Proceedings of the 2015 Conference on Empirical Methods in
  Natural Language Processing}, pages 190--200.

\bibitem[{Ren et~al.(2020)Ren, Guo, Lu, Zhou, Liu, Tang, Sundaresan, Zhou,
  Blanco, and Ma}]{ren2020codebleu}
Shuo Ren, Daya Guo, Shuai Lu, Long Zhou, Shujie Liu, Duyu Tang, Neel
  Sundaresan, Ming Zhou, Ambrosio Blanco, and Shuai Ma. 2020.
\newblock Codebleu: a method for automatic evaluation of code synthesis.
\newblock \emph{arXiv preprint arXiv:2009.10297}.

\bibitem[{Roncero and de~Almeida(2015)}]{roncero2015semantic}
Carlos Roncero and Roberto~G de~Almeida. 2015.
\newblock Semantic properties, aptness, familiarity, conventionality, and
  interpretive diversity scores for 84 metaphors and similes.
\newblock \emph{Behavior research methods}, 47(3):800--812.

\bibitem[{Sai et~al.(2022)Sai, Mohankumar, and Khapra}]{sai2022survey}
Ananya~B Sai, Akash~Kumar Mohankumar, and Mitesh~M Khapra. 2022.
\newblock A survey of evaluation metrics used for nlg systems.
\newblock \emph{ACM Computing Surveys (CSUR)}, 55(2):1--39.

\bibitem[{Stowe et~al.(2020)Stowe, Ribeiro, and Gurevych}]{stowe2020metaphoric}
Kevin Stowe, Leonardo Ribeiro, and Iryna Gurevych. 2020.
\newblock Metaphoric paraphrase generation.
\newblock \emph{arXiv preprint arXiv:2002.12854}.

\bibitem[{Tao et~al.(2018)Tao, Mou, Zhao, and Yan}]{tao2018ruber}
Chongyang Tao, Lili Mou, Dongyan Zhao, and Rui Yan. 2018.
\newblock Ruber: An unsupervised method for automatic evaluation of open-domain
  dialog systems.
\newblock In \emph{Thirty-Second AAAI Conference on Artificial Intelligence}.

\bibitem[{Tartakovsky and Shen(2018)}]{tartakovsky2018simple}
Roi Tartakovsky and Yeshayahu Shen. 2018.
\newblock ‘simple as a fire’: Making sense of the non-standard poetic
  simile.
\newblock \emph{Journal of Literary Semantics}, 47(2):103--119.

\bibitem[{Tevet and Berant(2021)}]{tevet2021evaluating}
Guy Tevet and Jonathan Berant. 2021.
\newblock Evaluating the evaluation of diversity in natural language
  generation.
\newblock In \emph{Proceedings of the 16th Conference of the European Chapter
  of the Association for Computational Linguistics: Main Volume}, pages
  326--346.

\bibitem[{Tversky(1977)}]{tversky1977features}
Amos Tversky. 1977.
\newblock Features of similarity.
\newblock \emph{Psychological review}, 84(4):327.

\bibitem[{Wang et~al.(2018)Wang, Singh, Michael, Hill, Levy, and
  Bowman}]{wang2018glue}
Alex Wang, Amanpreet Singh, Julian Michael, Felix Hill, Omer Levy, and Samuel
  Bowman. 2018.
\newblock Glue: A multi-task benchmark and analysis platform for natural
  language understanding.
\newblock In \emph{Proceedings of the 2018 EMNLP Workshop BlackboxNLP:
  Analyzing and Interpreting Neural Networks for NLP}, pages 353--355.

\bibitem[{Xiao et~al.(2016)Xiao, Alnajjar, Granroth-Wilding, Agres, Toivonen
  et~al.}]{xiao2016meta4meaning}
Ping Xiao, Khalid Alnajjar, Mark Granroth-Wilding, Kat Agres, Hannu Toivonen,
  et~al. 2016.
\newblock Meta4meaning: Automatic metaphor interpretation using corpus-derived
  word associations.
\newblock In \emph{Proceedings of the Seventh International Conference on
  Computational Creativity}. Sony CSL Paris.

\bibitem[{Zhang et~al.(2021)Zhang, Cui, Xia, Guo, Li, Wei, and
  Cui}]{zhang2021writing}
Jiayi Zhang, Zhi Cui, Xiaoqiang Xia, Yalong Guo, Yanran Li, Chen Wei, and
  Jianwei Cui. 2021.
\newblock Writing polishment with simile: Task, dataset and a neural approach.
\newblock In \emph{Proceedings of the AAAI Conference on Artificial
  Intelligence}, volume~35, pages 14383--14392.

\bibitem[{Zhang et~al.(2019)Zhang, Kishore, Wu, Weinberger, and
  Artzi}]{zhang2019bertscore}
Tianyi Zhang, Varsha Kishore, Felix Wu, Kilian~Q Weinberger, and Yoav Artzi.
  2019.
\newblock Bertscore: Evaluating text generation with bert.
\newblock \emph{arXiv preprint arXiv:1904.09675}.

\bibitem[{Zhang and Wan(2021)}]{zhang2021mover}
Yunxiang Zhang and Xiaojun Wan. 2021.
\newblock Mover: Mask, over-generate and rank for hyperbole generation.
\newblock \emph{arXiv preprint arXiv:2109.07726}.

\bibitem[{Zhao et~al.(2019)Zhao, Peyrard, Liu, Gao, Meyer, and
  Eger}]{zhao2019moverscore}
Wei Zhao, Maxime Peyrard, Fei Liu, Yang Gao, Christian~M Meyer, and Steffen
  Eger. 2019.
\newblock Moverscore: Text generation evaluating with contextualized embeddings
  and earth mover distance.
\newblock In \emph{Proceedings of the 2019 Conference on Empirical Methods in
  Natural Language Processing and the 9th International Joint Conference on
  Natural Language Processing (EMNLP-IJCNLP)}, pages 563--578.

\bibitem[{Zheng et~al.(2019)Zheng, Song, Hu, Fu, and Zhou}]{zheng2019love}
Danning Zheng, Ruihua Song, Tianran Hu, Hao Fu, and Jin Zhou. 2019.
\newblock “love is as complex as math”: Metaphor generation system for
  social chatbot.
\newblock In \emph{Workshop on Chinese Lexical Semantics}, pages 337--347.
  Springer.

\bibitem[{Zhu et~al.(2018)Zhu, Lu, Zheng, Guo, Zhang, Wang, and
  Yu}]{zhu2018texygen}
Yaoming Zhu, Sidi Lu, Lei Zheng, Jiaxian Guo, Weinan Zhang, Jun Wang, and Yong
  Yu. 2018.
\newblock Texygen: A benchmarking platform for text generation models.
\newblock In \emph{The 41st International ACM SIGIR Conference on Research \&
  Development in Information Retrieval}, pages 1097--1100.

\end{thebibliography}
\bibliographystyle{acl_natbib}

\clearpage

\appendix

\section{Human Ratings} \label{sec:human_ratings}
The instructions given to raters are detailed as follows:
\begin{enumerate}
    \item All raters are provided with the necessary background information on similes and the simile generation task, including the definition of similes, the main simile components, and the motivation of our proposed criteria.
    \item To ensure the quality of ratings, all the raters label a small set of 20 samples to reach an agreement on the labeling criteria for each metric before the formal labeling.
    \item For each perspective (i.e. quality, creativity, informativeness), three raters are asked to rate each simile from 1 to 5, where 1 denotes the worst and 5 denotes the best. \textbf{The examples of our human ratings} are provided in Tab.~\ref{tab:ratings_exp}.
    \item During the rating, raters are asked to specifically label the simile-literal sentence pairs which lack context and are thus difficult to rate (e.g. ``\textit{Nobody can shoot.}'').
\end{enumerate}

\section{NDCG Formulation}  \label{sec:appendix_ndcg}
In our setting, the optimal rankings are human rankings.
Hence, given $m$ simile candidates $\mathcal{S} = \{s_1, s_2, ..., s_m\}$, the NDCG@k given by each automatic metric is defined as follows:


\vspace{-2mm}
\begin{small}
\begin{align}
		& \text{NDCG}(k) = \frac{\text{DCG}(\mathcal{O}_{\text{hypo}}, k)}{\text{DCG}(\mathcal{O}_{\text{ref}}, k)} \\
            & \text{DCG}(\mathcal{O}, k) = \sum_{i=1}^k \frac{{\mathcal{O}[\mathcal{I}(i)]} }{log_2(1+i)} 
\end{align} 
\end{small}
where $O_{\text{ref}}$ and $O_{\text{hypo}}$ represent the score list given by humans and each automatic metric respectively, $\mathcal{O}[j]$ denote the score of $s_j$,  $\mathcal{I}(i)$ denotes the index of the $i$-th largest score in $\mathcal{O}$.

\section{The Implementation of Prior Metrics} \label{sec:appendix_evaluation}
We report the packages used to implement prior automatic metrics in Tab.~\ref{tab:packages}.
For the metric denoted with an asterisk(*), we apply the corresponding package to implement the key parts, based on the definition from the cited papers.
The formulation of NDCG in our setting is provided in Appx. \ref{sec:appendix_ndcg}.
The rest of the metrics are entirely implemented by us according to the cited papers.

\begin{table}[!htbp]
\small
\centering
\begin{tabular}{ll}
  \toprule  
\textbf{Metric}           & \textbf{Packages} \\
\midrule
BLEU, METEOR      & NLTK           \\
Rouge             & rouge          \\
BERTScore         & bert\_score    \\
Self-BLEU*        & NLTK           \\
Distinct n-grams* & NLTK    \\
  \bottomrule  
\end{tabular}
        \caption{The packages used to implement the metrics.}    
        \label{tab:packages}
\end{table}

\section{The Details of MAPS-KB} \label{sec:mapskb}
MAPS-KB~\cite{he2022maps} is a million-scale probabilistic simile knowledge, containing 4.3 million simile triplets from 70 GB corpora, along with frequency and two probabilistic metrics, \textit{plausibility} and \textit{typicality}, to model each triplet.
The simile triplet is in the form of (topic, property, vehicle)$(t, p, v)$.

In our paper, we specifically adopt the \textit{frequency} and \textit{plausibility} information from MAPS-KB to implement our relevance metric. 
With regard to \textit{plausibility}, it evaluates the quality of simile triplets based on the confidence score of their supporting simile instances (simile sentence, topic, property, vehicle)$(s_i, t, p, v)$.
In each simile instance, the topic and vehicle are extracted from the simile sentence, while the property is generated via generative commonsense model COMET~\cite{bosselut2019comet} and prompting the PLMs.
MAPS-KB adopt the \textit{noisy-or} model to measure the plausibility of the triplet $(t, p, v)$, which is defined as follows:

\vspace{-3mm}
\begin{footnotesize}
\begin{gather*}
       \mathcal{P}(t,p,v) = 1 - \prod_{i = 1}^{\eta}(1 - S(s_i, t, p, v)),
\end{gather*}
\end{footnotesize}
where $S(s_i, t, p, v)=P(p|s_i, t, v)$ is the confidence score of each simile instance during generation and $\eta$ is the number of simile instances supporting the simile triplet ($t$, $p$, $v$).

\begin{table*}[t]
\resizebox{\textwidth}{!}{
\begin{tabular}{|c|l|c|c|c|c|}
\hline
\textbf{\# }                & \multicolumn{1}{c|}{\textbf{Literal Sentence}}                                                                                                                                                                                                                                    & \textbf{Vehicles in the Generated Similes}                        & \textbf{Q}   & \textbf{C}   & \textbf{I}   \\ \hline
\multirow{5}{*}{\textbf{1}} & \multirow{5}{*}{Some raindrops struck the roof, window and ran down its panes \sethlcolor{yellow}\hl{{[}insert{]}}.}                                                                                                                                               & like diamonds                            & 2.3 & 3.3 & 2.0 \\ \cline{3-6} 
                   &                                                                                                                                                                                                                                            & like tears                               & 3.3 & 3.3 & 2.0 \\ \cline{3-6} 
                   &                                                                                                                                                                                                                                            & like arrows                              & 1.0 & 3.0 & 2.0 \\ \cline{3-6} 
                   &                                                                                                                                                                                                                                            & like a stream                            & \textbf{4.0} & 2.7 & 2.3 \\ \cline{3-6} 
                   &                                                                                                                                                                                                                                            & like a stream of diamonds                & \textbf{4.0} & \textbf{4.7} & \textbf{4.0} \\ \hline
\multirow{5}{*}{\textbf{2}} & \multirow{4}{*}{As suddenly as she'd jumped up from the sofa, Jaklin collapsed \sethlcolor{yellow}\hl{{[}insert{]}}.}                                                                                                                                             & like a rag doll                          & 3.0 & 3.3 & 2.7 \\ \cline{3-6} 
                   &                                                                                                                                                                                                                                            & like a deflated balloon                  & \textbf{4.7} & 4.0 & \textbf{3.3} \\ \cline{3-6} 
                   &                                                                                                                                                                                                                                            & like a pricked bladder                   & 3.0 & \textbf{4.7} & \textbf{3.3} \\ \cline{3-6} 
                   &                                                                                                                                                                                                                                            & like a pricked balloon                   & 4.3 & 4.3 & \textbf{3.3} \\ \cline{2-6} 
                   & As suddenly as \sethlcolor{yellow}\hl{{[}insert{]}} she'd jumped up from the sofa, Jaklin collapsed.                                                                                                                                                              & like a flash                             & 3.3 & 2.0 & 2.3 \\ \hline
\multirow{5}{*}{\textbf{3}} & \multirow{5}{*}{In the other direction the Empire State Building loomed \sethlcolor{yellow}\hl{{[}insert{]}}.}                                                                                                                                                     & like a dark shadow                       & 4.0 & 2.3 & 3.3 \\ \cline{3-6} 
                   &                                                                                                                                                                                                                                            & like a huge black monster                & \textbf{4.7} & 3.3 & 4.0 \\ \cline{3-6} 
                   &                                                                                                                                                                                                                                            & like a giant black monster               & \textbf{4.7} & 3.7 & 4.0 \\ \cline{3-6} 
                   &                                                                                                                                                     & like a huge black shadow                 & 4.3 & 3.0 & 4.0 \\ \cline{3-6} 
                   &                                                                                                                                                                                                                                            & like a huge black monster of destruction & \textbf{4.7} & \textbf{4.3} & \textbf{5.0} \\ \hline
\multirow{5}{*}{\textbf{4}} & \multirow{5}{*}{\begin{tabular}[c]{@{}l@{}}His hormones boiled and steamed \sethlcolor{yellow}\hl{{[}insert{]}} and yet he did not reach for \\ the succulent young flesh there beside him.\end{tabular} }                                                                                                    & like a boiling caldron                   & 3.0 & 4.3 & 3.0 \\ \cline{3-6} 
                   &                                                                                                                                                                                                                                            & like a volcano                           & 4.3 & 2.7 & 2.0 \\ \cline{3-6} 
                   &                                                                                                                                                                                                                                            & like a boiling cauldron                  & 4.0 & \textbf{4.7} & 3.0 \\ \cline{3-6} 
                   &                                                                                                                                                                                                                                            & like a cauldron of boiling water         & \textbf{4.7} & \textbf{4.7} & \textbf{4.7} \\ \cline{3-6} 
                   &                                                                                                                                                                                                                                            & like a cauldron of boiling water*        & \textbf{4.7} & \textbf{4.7} & \textbf{4.7} \\ \hline
\multirow{5}{*}{\textbf{5}} & \multirow{5}{*}{\begin{tabular}[c]{@{}l@{}}The coil whistled through the air. It fell right over the mate's shoulder. \\  He clutched at it as the fore, topmast crosstrees, with the full force of the \\ surge, struck him from behind, and he sank \sethlcolor{yellow}\hl{{[}insert{]}}.\end{tabular} 
 } & like a stone                             & \textbf{4.7} & 1.7 & 2.0 \\ \cline{3-6} 
                   &                                                                                                                                                                                                                                            & like a log                               & 3.0 & 1.7 & 2.0 \\ \cline{3-6} 
                   &                                                                                                                                                                                                                                            & like lead                                & 4.3 & 2.3 & 1.7 \\ \cline{3-6} 
                   &                                                                                                                                                                                                                                            & like an empty sack                       & 1.3 & \textbf{3.7} & \textbf{3.0} \\ \cline{3-6} 
                   &                                                                                                                                                                                                                                            & like an empty barrel                     & 1.3 & \textbf{3.7} & \textbf{3.0} \\ \hline
\end{tabular}
}

    \caption{Examples of human ratings for each perspective (Q, C, I denoting \textit{Quality}, \textit{Creativity}, \textit{Informativeness}, respectively). 
    The indicators ``[insert]'' denotes the insertion positions of vehicles within the generated similes given by models, which do not exist in the literal sentences.
    Bold numbers indicate the highest ranking among the simile candidates generated from a literal sentence. An asterisk (*) indicates that the generated simile introduces noise to the context word through additions, deletions, or changes within two words.}
    \label{tab:ratings_exp}
\end{table*}

\begin{table*}[!htbp]
\centering

\resizebox{\textwidth}{!}{
\begin{tabular}{|c|c|l|lll|}
\hline
\multirow{2}{*}{\textbf{\#}} &
  \multirow{2}{*}{\textbf{Method}} &
  \multicolumn{1}{c|}{\multirow{2}{*}{\textbf{Literal Sentence}}} &
  \multicolumn{3}{c|}{\textbf{Vehicles in the Generated Similes}} \\ \cline{4-6} 
 &
   &
  \multicolumn{1}{c|}{} &
  \multicolumn{1}{c|}{\textbf{Original Rank}} &
  \multicolumn{1}{c|}{\textbf{HAUSER Rank}} &
  \multicolumn{1}{c|}{\textbf{Human Rank}} \\ \hline

\multirow{10}{*}{\textbf{1}} &
  \multirow{5}{*}{BM25} &
  \multirow{5}{*}{\begin{tabular}[c]{@{}l@{}}Stefan moved \sethlcolor{yellow}\hl{{[}Insert{]}}, every movement easy \\ and precisely controlled.\end{tabular}} &
  \multicolumn{1}{l|}{\cellcolor{green4}like water} &
  \multicolumn{1}{l|}{\textbf{\cellcolor{green1}like a ballerina}} &
  \textbf{\cellcolor{green1}like a ballerina} \\
 &
   &
   &
  \multicolumn{1}{l|}{\cellcolor{green5}like hell} &
  \multicolumn{1}{l|}{\cellcolor{green2}like a predator} &
  \cellcolor{green2}like a predator \\
 &
   &
   &
  \multicolumn{1}{l|}{\textbf{\cellcolor{green1}like a ballerina}} &
  \multicolumn{1}{l|}{\cellcolor{green3}like a drum} &
  \cellcolor{green3}like a drum \\
 &
   &
   &
  \multicolumn{1}{l|}{\cellcolor{green3}like a drum} &
  \multicolumn{1}{l|}{\cellcolor{green4}like water} &
  \cellcolor{green4}like water \\
 &
   &
   &
  \multicolumn{1}{l|}{\cellcolor{green2}like a predator} &
  \multicolumn{1}{l|}{\cellcolor{green5}like hell} &
 \cellcolor{green5}like hell \\ \cline{2-6} 
 &
  \multirow{5}{*}{Ours} &
  \multirow{5}{*}{\begin{tabular}[c]{@{}l@{}}Stefan moved \hl{{[}Insert{]}}, every movement easy \\ and precisely controlled.\end{tabular}} &
  \multicolumn{1}{l|}{\cellcolor{green3}like a cat} &
  \multicolumn{1}{l|}{\textbf{\cellcolor{green1}like a dancer}} &
  \textbf{\cellcolor{green1}like a dancer}\\
 &
   &
   &
  \multicolumn{1}{l|}{\textbf{\cellcolor{green1}like a dancer}} &
  \multicolumn{1}{l|}{\cellcolor{green2}like an automaton} &
  \cellcolor{green2}like an automaton   \\
 &
   &
   &
  \multicolumn{1}{l|}{\cellcolor{green5}like lightning} &
  \multicolumn{1}{l|}{\cellcolor{green5}like lightning} &
  \cellcolor{green3}like a cat \\
 &
   &
   &
  \multicolumn{1}{l|}{\cellcolor{green2}like an automaton} &
  \multicolumn{1}{l|}{\cellcolor{green3}like a cat} &
  \cellcolor{green4}like a cat* \\
 &
   &
   &
  \multicolumn{1}{l|}{\cellcolor{green4}like a cat*} &
  \multicolumn{1}{l|}{\cellcolor{green4}like a cat*} &
 \cellcolor{green5}like lightning \\ \hline

\multirow{10}{*}{\textbf{2}} &
  \multirow{5}{*}{BM25} &
  \multirow{5}{*}{But his next line called for him to howl \hl{{[}Insert{]}}.} &
  \multicolumn{1}{l|}{\cellcolor{green2}like a fiend} &
  \multicolumn{1}{l|}{\textbf{\cellcolor{green1}like a wounded buffalo}} &
  \textbf{\cellcolor{green1}like a wounded buffalo} \\
 &
   &
   &
  \multicolumn{1}{l|}{\cellcolor{green5}like a drug} &
  \multicolumn{1}{l|}{\cellcolor{green2}like a fiend} &
   \cellcolor{green2}like a fiend\\
 &
   &
   &
  \multicolumn{1}{l|}{\cellcolor{green3}like a chicken} &
  \multicolumn{1}{l|}{\cellcolor{green4}like a trail} &
  \cellcolor{green3}like a chicken \\
 &
   &
   &
  \multicolumn{1}{l|}{\cellcolor{green4}like a trail} &
  \multicolumn{1}{l|}{\cellcolor{green3}like a chicken} &
  \cellcolor{green4}like a trail \\
 &
   &
   &
  \multicolumn{1}{l|}{\textbf{\cellcolor{green1}like a wounded buffalo}} &
  \multicolumn{1}{l|}{\cellcolor{green5}like a drug} &
  \cellcolor{green5}like a drug \\ \cline{2-6} 
 &
  \multirow{5}{*}{Ours} &
  \multirow{5}{*}{But his next line called for him to howl \hl{{[}Insert{]}}.} &
  \multicolumn{1}{l|}{\cellcolor{green4}like a wolf} &
  \multicolumn{1}{l|}{\textbf{\cellcolor{green1}like a wounded animal}} &
  \textbf{\cellcolor{green1}like a wounded animal} \\
 &
   &
   &
  \multicolumn{1}{l|}{\cellcolor{green5}like a dog} &
  \multicolumn{1}{l|}{\cellcolor{green5}like a dog} &
  \cellcolor{green2}like a coyote \\
 &
   &
   &
  \multicolumn{1}{l|}{\cellcolor{green2}like a coyote} &
  \multicolumn{1}{l|}{\cellcolor{green2}like a coyote} &
\cellcolor{green3}like a coyote* \\
 &
   &
   &
  \multicolumn{1}{l|}{\textbf{\cellcolor{green1}like a wounded animal.}} &
  \multicolumn{1}{l|}{\cellcolor{green3}like a coyote*} &
   \cellcolor{green4}like a wolf \\
 &
   &
   &
  \multicolumn{1}{l|}{\cellcolor{green3}like a coyote*} &
  \multicolumn{1}{l|}{\cellcolor{green4}like a wolf} &
 \cellcolor{green5}like a dog \\ \hline

\multirow{10}{*}{\textbf{3}} &
  \multirow{5}{*}{BM25} &
  \multirow{5}{*}{\begin{tabular}[c]{@{}l@{}}She wondered absently if those soldiers would \\ survive the coming war, if they would earn \\ glory or run \hl{{[}Insert{]}}.\end{tabular}} &
  \multicolumn{1}{l|}{\cellcolor{green3}like a rabbit} &
  \multicolumn{1}{l|}{\textbf{\cellcolor{green1}like a very coward}} &
  \textbf{\cellcolor{green1}like a very coward} \\
 &
   &
   &
  \multicolumn{1}{l|}{\cellcolor{green5}like bees about their friend} &
  \multicolumn{1}{l|}{\cellcolor{green5}like bees about their friend} &
  \cellcolor{green2}like a pack of wolves \\
 &
   &
   &
  \multicolumn{1}{l|}{\cellcolor{green4}like wildfire} &
  \multicolumn{1}{l|}{\cellcolor{green2}like a pack of wolves} &
 \cellcolor{green3}like a rabbit \\
 &
   &
   &
  \multicolumn{1}{l|}{\textbf{\cellcolor{green1}like a very coward}} &
  \multicolumn{1}{l|}{\cellcolor{green4}like wildfire} &
  \cellcolor{green4}like wildfire  \\
 &
   &
   &
  \multicolumn{1}{l|}{\cellcolor{green2}like a pack of wolves} &
  \multicolumn{1}{l|}{\cellcolor{green3}like a rabbit} &
  \cellcolor{green5}like bees about their friend \\ \cline{2-6} 
 &
  \multirow{5}{*}{Ours} &
  \multirow{5}{*}{\begin{tabular}[c]{@{}l@{}}She wondered absently if those soldiers would \\ survive the coming war, if they would earn \\ glory or run \hl{{[}Insert{]}}.\end{tabular}} &
  \multicolumn{1}{l|}{\cellcolor{green3}like cowards} &
  \multicolumn{1}{l|}{\cellcolor{green2}like scared rabbits} &
  \textbf{\cellcolor{green1}like frightened sheep} \\
 &
   &
   &
  \multicolumn{1}{l|}{\cellcolor{green2}like scared rabbits} &
  \multicolumn{1}{l|}{\cellcolor{green5}like hares} &
   \cellcolor{green2}like scared rabbits \\
 &
   &
   &
  \multicolumn{1}{l|}{\textbf{\cellcolor{green1}like frightened sheep}} &
  \multicolumn{1}{l|}{\textbf{\cellcolor{green1}like frightened sheep}} &
  \cellcolor{green3}like cowards \\
 &
   &
   &
  \multicolumn{1}{l|}{\cellcolor{green5}like hares} &
  \multicolumn{1}{l|}{\cellcolor{green3}like cowards} &
  \cellcolor{green4}like cowards*  \\
 &
   &
   &
  \multicolumn{1}{l|}{\cellcolor{green4}like cowards*} &
  \multicolumn{1}{l|}{\cellcolor{green4}like cowards*} &
  \cellcolor{green5}like hares \\ \hline

\multirow{10}{*}{\textbf{4}} &
  \multirow{5}{*}{BM25} &
  \multirow{5}{*}{\begin{tabular}[c]{@{}l@{}}As suddenly as she'd jumped up from the sofa, \\ Jaklin collapsed \hl{{[}Insert{]}}.\end{tabular}} &
  \multicolumn{1}{l|}{\textbf{\cellcolor{green1}like a pricked bubble}} &
  \multicolumn{1}{l|}{\cellcolor{green3}like a grocery bag} &
  \textbf{\cellcolor{green1}like a pricked bubble} \\
 &
   &
   &
  \multicolumn{1}{l|}{\cellcolor{green5}like a boy} &
  \multicolumn{1}{l|}{\textbf{\cellcolor{green1}like a pricked bubble}} &
  \cellcolor{green2}like a ragdoll \\
 &
   &
   &
  \multicolumn{1}{l|}{\cellcolor{green4}like a panther} &
  \multicolumn{1}{l|}{\cellcolor{green2}like a ragdoll} &
  \cellcolor{green3}like a grocery bag \\
 &
   &
   &
  \multicolumn{1}{l|}{\cellcolor{green2}like a ragdoll} &
  \multicolumn{1}{l|}{\cellcolor{green5}like a boy} &
  \cellcolor{green4}like a panther \\
 &
   &
   &
  \multicolumn{1}{l|}{\cellcolor{green3}like a grocery bag} &
  \multicolumn{1}{l|}{\cellcolor{green4}like a panther} &
  \cellcolor{green5}like a boy \\ \cline{2-6} 
 &
  \multirow{5}{*}{Ours} &
  \multirow{5}{*}{\begin{tabular}[c]{@{}l@{}}As suddenly as she'd jumped up from the sofa, \\ Jaklin collapsed \hl{{[}Insert{]}}.\end{tabular}} &
  \multicolumn{1}{l|}{\cellcolor{green3}like a rag doll} &
  \multicolumn{1}{l|}{\cellcolor{green5}like a sack of potatoes*} &
  \textbf{\cellcolor{green1}like a deflated balloon} \\
 &
   &
   &
  \multicolumn{1}{l|}{\textbf{\cellcolor{green1}like a deflated balloon}} &
  \multicolumn{1}{l|}{\textbf{\cellcolor{green1}like a deflated balloon}} &
  \cellcolor{green2}like a pricked balloon \\
 &
   &
   &
  \multicolumn{1}{l|}{\cellcolor{green4}like a sack of potatoes} &
  \multicolumn{1}{l|}{\cellcolor{green2}like a pricked balloon} &
  \cellcolor{green3}like a rag doll \\
 &
   &
   &
  \multicolumn{1}{l|}{\cellcolor{green2}like a pricked balloon} &
  \multicolumn{1}{l|}{\cellcolor{green4}like a sack of potatoes} &
  \cellcolor{green4}like a sack of potatoes \\
 &
   &
   &
  \multicolumn{1}{l|}{\cellcolor{green5}like a sack of potatoes*} &
  \multicolumn{1}{l|}{\cellcolor{green3}like a rag doll} &
  \cellcolor{green5}like a sack of potatoes* \\ \hline

\multirow{10}{*}{\textbf{5}} &
  \multirow{5}{*}{BM25} &
  \multirow{5}{*}{They gleamed \hl{{[}Insert{]}}.} &
  \multicolumn{1}{l|}{\cellcolor{green2}like golden fire} &
  \multicolumn{1}{l|}{\textbf{\cellcolor{green1}like the eyes of great cats}} &
   \textbf{\cellcolor{green1}like the eyes of great cats}  \\
 &
   &
   &
  \multicolumn{1}{l|}{\cellcolor{green4}like silver} &
  \multicolumn{1}{l|}{\cellcolor{green2}like golden fire} &
  \cellcolor{green2}like golden fire  \\
 &
   &
   &
  \multicolumn{1}{l|}{\textbf{\cellcolor{green1}like the eyes of great cats}} &
  \multicolumn{1}{l|}{\cellcolor{green4}like silver} &
  \cellcolor{green3}like sparks of fire  \\
 &
   &
   &
  \multicolumn{1}{l|}{\cellcolor{green5}like a second skin} &
  \multicolumn{1}{l|}{\cellcolor{green3}like sparks of fire} &
  \cellcolor{green4}like silver\\
 &
   &
   &
  \multicolumn{1}{l|}{\cellcolor{green3}like sparks of fire} &
  \multicolumn{1}{l|}{\cellcolor{green5}like a second skin} &
  \cellcolor{green5}like a second skin \\ \cline{2-6} 
 &
  \multirow{5}{*}{Ours} &
  \multirow{5}{*}{They gleamed \hl{{[}Insert{]}}.} &
  \multicolumn{1}{l|}{\cellcolor{green4}like polished ebony} &
  \multicolumn{1}{l|}{\cellcolor{green3}like the eyes of a cat} &
    \textbf{\cellcolor{green1}like the eyes of a wild beast} \\
 &
   &
   &
  \multicolumn{1}{l|}{\cellcolor{green5}like polished steel} &
  \multicolumn{1}{l|}{\cellcolor{green2}like the eyes of a wild animal} &
  \cellcolor{green2}like the eyes of a wild animal  \\
 &
   &
   &
  \multicolumn{1}{l|}{\cellcolor{green3}like the eyes of a cat} &
  \multicolumn{1}{l|}{\textbf{\cellcolor{green1}like the eyes of a wild beast}} &
  \cellcolor{green3}like the eyes of a cat \\
 &
   &
   &
  \multicolumn{1}{l|}{\cellcolor{green2}like the eyes of a wild animal} &
  \multicolumn{1}{l|}{\cellcolor{green4}like polished ebony} &
  \cellcolor{green4}like polished ebony \\
 &
   &
   &
  \multicolumn{1}{l|}{\textbf{\cellcolor{green1}like the eyes of a wild beast}} &
  \multicolumn{1}{l|}{\cellcolor{green5}like polished steel} &
  \cellcolor{green5}like polished steel \\ \hline


\multirow{10}{*}{\textbf{6}} &
  {\multirow{5}{*}{BM25}} &
  \multirow{5}{*}{The idea resounded \hl{{[}Insert{]}} throughout the land.} &

  \multicolumn{1}{l|}{\cellcolor{green3}like a gong} &
  \multicolumn{1}{l|}{\cellcolor{green4}like the beating of a bass drum} &
 \multicolumn{1}{l|}{\begin{tabular}[c]{@{}l@{}}\textbf{\cellcolor{green1}like the crack of a whip in the} \\ \textbf{\cellcolor{green1}silence of the hall}\end{tabular}} \\

 &
  \multicolumn{1}{l|}{} &
   &
  \multicolumn{1}{l|}{\cellcolor{green5}like an agonized lament} &
\multicolumn{1}{l|}{\begin{tabular}[c]{@{}l@{}}\textbf{\cellcolor{green1}like the crack of a whip in the} \\ \textbf{\cellcolor{green1}silence of the hall}\end{tabular}} &
   \cellcolor{green2}like prolonged theater applause\\
 &

  \multicolumn{1}{l|}{} &
   &
  \multicolumn{1}{l|}{\cellcolor{green4}like the beating of a bass drum} &
  \multicolumn{1}{l|}{\cellcolor{green2}like prolonged theater applause} &
  \cellcolor{green3}like a gong \\
 &

  \multicolumn{1}{l|}{} &
   &
\multicolumn{1}{l|}{\begin{tabular}[c]{@{}l@{}}\textbf{\cellcolor{green1}like the crack of a whip in the} \\ \textbf{\cellcolor{green1}silence of the hall}\end{tabular}} &
  \multicolumn{1}{l|}{\cellcolor{green5}like an agonized lament} &
  \cellcolor{green4}like the beating of a bass drum\\
 &

  \multicolumn{1}{l|}{} &
   &
  \multicolumn{1}{l|}{\cellcolor{green2}like prolonged theater applause} &
  \multicolumn{1}{l|}{\cellcolor{green3}like a gong} &
  \cellcolor{green5}like an agonized lament \\ \cline{2-6} 
 &
  {\multirow{5}{*}{Ours}} &
  \multirow{5}{*}{The idea resounded \hl{{[}Insert{]}} throughout the land.} &
  \multicolumn{1}{l|}{\cellcolor{green3}like thunder} &
  \multicolumn{1}{l|}{\cellcolor{green4}like a trumpet} &
  \textbf{\cellcolor{green1}like a thunderclap} \\
 &
  \multicolumn{1}{l|}{} &
   &
  \multicolumn{1}{l|}{\textbf{\cellcolor{green1}like a thunderclap}} &
  \multicolumn{1}{l|}{\cellcolor{green2}like a thunderclap*} &
  \cellcolor{green2}like a thunderclap* \\
 &
  \multicolumn{1}{l|}{} &
   &
  \multicolumn{1}{l|}{\cellcolor{green5}like an earthquake} &
  \multicolumn{1}{l|}{\textbf{\cellcolor{green1}like a thunderclap}} &
   \cellcolor{green3}like thunder\\
 &
  \multicolumn{1}{l|}{} &
   &
  \multicolumn{1}{l|}{\cellcolor{green4}like a trumpet} &
  \multicolumn{1}{l|}{\cellcolor{green5}like an earthquake} &
   \cellcolor{green4}like a trumpet \\
 &
  \multicolumn{1}{l|}{} &
   &
  \multicolumn{1}{l|}{\cellcolor{green2}like a thunderclap*} &
  \multicolumn{1}{l|}{\cellcolor{green3}like thunder} &
  \cellcolor{green5}like an earthquake \\ \hline
\end{tabular}
}
\caption{ The examples of simile candidates reranked via \texttt{HAUSER}, which are generated by various methods. 
The indicators ``[insert]'' denotes the insertion positions of vehicles within the generated similes given by models, which do not exist in the literal sentences.
An asterisk (*) indicates that the generated simile introduces noise to the context word through additions, deletions, or changes within two words. 
A darker shade of green indicates a higher rank bestowed by humans.
}
\label{tab:application}
\end{table*}

\end{document}